\documentclass{article}
\usepackage[dvipsnames]{xcolor}

\usepackage[final,nonatbib]{neurips_2022}

\usepackage{amsmath,amsfonts,bm}

\def\eqref#1{equation~\ref{#1}}
\def\1{\bm{1}}

\def\vh{{\bm{h}}}

\def\vo{{\bm{o}}}

\def\vu{{\bm{u}}}

\def\vx{{\bm{x}}}

\def\mW{{\bm{W}}}
\def\mX{{\bm{X}}}

\DeclareMathAlphabet{\mathsfit}{\encodingdefault}{\sfdefault}{m}{sl}
\SetMathAlphabet{\mathsfit}{bold}{\encodingdefault}{\sfdefault}{bx}{n}

\usepackage[utf8]{inputenc} % allow utf-8 input
\usepackage[T1]{fontenc}    % use 8-bit T1 fonts
\usepackage{hyperref}       % hyperlinks
\usepackage{url}            % simple URL typesetting
\usepackage{enumitem}
\usepackage{wrapfig}
\usepackage{booktabs}       % professional-quality tables
\usepackage{amsfonts}       % blackboard math symbols
\usepackage{amsmath, amsthm, amssymb}
\usepackage{nicefrac}       % compact symbols for 1/2, etc.
\usepackage{microtype}      % microtypography
\usepackage{xcolor}         % colors
\usepackage{adjustbox}
\usepackage{multirow}
\usepackage{textcomp}
\usepackage{pifont}
\usepackage[font=small]{caption}
\usepackage{subcaption}

\usepackage[misc]{ifsym}
\usepackage{floatrow}

\hypersetup{colorlinks, breaklinks, citecolor=[rgb]{0.0007, 0.44, 0.737},%{0,0.5411,0.45},%et al
anchorcolor=[rgb]{0.0007, 0.44, 0.737}, linkcolor=[rgb]{0.0007, 0.44, 0.737},%[rgb]{0.74,0.05,0.},% figure & section
urlcolor=[rgb]{0.0007, 0.44, 0.737} % url color
}

\theoremstyle{plain}

\theoremstyle{definition}

\def\equationautorefname~#1\null{Eq.~(#1)\null}

\newcommand{\aref}[1]{\hyperref[#1]{Appendix~\ref{#1}}} % defined for sections in Appendix!
\makeatletter
\newif\if@restonecol
\makeatother

\usepackage[ruled,vlined]{algorithm2e}%[ruled,vlined]{
\usepackage{algpseudocode}

\title{Outlier Suppression: Pushing the Limit of Low-bit Transformer Language Models}

\author{Xiuying Wei\textsuperscript{1, 2}\ , Yunchen Zhang\textsuperscript{2, 4}\ , Xiangguo Zhang\textsuperscript{2}\ , Ruihao Gong\textsuperscript{1, 2}, \\
\textbf{Shanghang Zhang\textsuperscript{3}\ , Qi Zhang\textsuperscript{2}\ , Fengwei Yu\textsuperscript{2}\ , Xianglong Liu\textsuperscript{1}\thanks{Corresponding author.}}\\
\textsuperscript{1}State Key Lab of Software Development Environment, Beihang University\\
\textsuperscript{2}SenseTime Research,
\textsuperscript{3}Peking University\\
\textsuperscript{4}University of Electronic Science and Technology of China\\
\footnotesize{\texttt{\{weixiuying, zhangyunchen, zhangxiangguo, gongruihao\}@sensetime.com}}\\
\footnotesize{\texttt{
    shanghang@pku.edu.cn,
    xlliu@buaa.edu.cn}}
}

\begin{document}

\maketitle

\begin{abstract}
Transformer architecture has become the fundamental element of the widespread natural language processing~(NLP) models. With the trends of large NLP models, the increasing memory and computation costs hinder their efficient deployment on resource-limited devices. Therefore, transformer quantization attracts wide research interest. Recent work recognizes that structured outliers are the critical bottleneck for quantization performance. However, their proposed methods increase the computation overhead and still leave the outliers there. To fundamentally address this problem, this paper delves into the inherent inducement and importance of the outliers. We discover that $\bm \gamma$ in LayerNorm (LN) acts as a sinful amplifier for the outliers, and the importance of outliers varies greatly where some outliers provided by a few tokens cover a large area but can be clipped sharply without negative impacts. Motivated by these findings, we propose an outlier suppression framework including two components: Gamma Migration and Token-Wise Clipping. The Gamma Migration migrates the outlier amplifier to subsequent modules in an equivalent transformation, contributing to a more quantization-friendly model without any extra burden. The Token-Wise Clipping takes advantage of the large variance of token range and designs a token-wise coarse-to-fine pipeline, obtaining a clipping range with minimal final quantization loss in an efficient way. This framework effectively suppresses the outliers and can be used in a plug-and-play mode. Extensive experiments prove that our framework surpasses the existing works and, for the first time, pushes the 6-bit post-training BERT quantization to the full-precision (FP) level. Our code is available at \url{https://github.com/wimh966/outlier_suppression}.

\end{abstract}

\section{Introduction}
%Transformer-based models or bert-like models ???
Transformer \cite{vaswani2017attention} has been one of the most common architectures in natural language processing along with lots of popular self-supervised models, such as BERT \cite{devlin2018bert}, RoBERTa \cite{liu2019roberta}, XLNet \cite{yang2019xlnet} and BART \cite{lewis2019bart}. While these pre-trained models have demonstrated a significant superiority in performance, the memory and computation overheads have been a popular concern, particularly in the real development. Therefore, model compression~\cite{han2015deepcompress,hinton2015distilling,zoph2016nas,2021-iccv-oqa} has attracted much attention from both academia and industry.
Among them, quantization~\cite{2019-iccv-dsq,esser2019learned,bhalgat2020lsq+, 2020-cvpr-int8training,hubara2021accurate,nahshan2019lapq,cai2020zeroq,2021-cvpr-dsg,nagel2020adaround,2021-iclr-brecq,2022-iclr-qdrop}, working in the low-precision arithmetic fashion, is one of the key approaches to compress large models and fit them into the lightweight devices.

These days, researchers focus more on quantization of Transformer-based models. \cite{zafrir2019q8bert} proposes an 8-bit quantization scheme for BERT-like models. \cite{shen2020q} advises a group-wise quantization technique and analyzes mixed-precision using second-order Hessian information. \cite{zhang2020ternarybert, bai2020binarybert} combine distillation with quantization. \cite{kim2021bert} approximates nonlinear operations to implement integer-only quantization. Nonetheless, few studies investigate the inherent bottleneck of quantizing Transformer-based models.

% Recently, some papers~\cite{bondarenko2021understanding,kovaleva2021bert} indicate that there exist significantly large outliers (some close to 100) in NLP models than in the computer vision ones, bringing devastating damage to the quantization accuracy (e.g., a 12\% drop even for the 8-bit). Specifically, these extreme outliers behave in structured patterns, bringing devastating damage to the quantization accuracy (e.g., a 12\% drop even for the 8-bit). For this critical outlier problem, existing method \cite{bondarenko2021understanding} chooses bypassing solutions such as a finer quantization granularity. 
% However, this scheme causes an increased computation cost and unavoidably hinders the acceleration effect.

Recently, some papers~\cite{bondarenko2021understanding, luo2020positional} indicate that the Transformer-based models hold significantly large outliers (even close to 100) and these extreme outliers behave in structured patterns (mainly gather at a few embedding dimensions and even become larger on unique tokens). 
These special outliers can bring devastating damage to the quantization performance (e.g., a 12\% drop even for the 8-bit \cite{bondarenko2021understanding}). To combat this challenge, existing method~\cite{bondarenko2021understanding} chooses bypassing solutions such as a finer quantization granularity. However, this scheme causes an increased computation cost and unavoidably hinders the acceleration effect.

In this paper, to suppress the outliers rather than walk around them, we make an in-depth analysis to investigate the inducement of the outliers and the impact of clipping the outliers. For the inducement, we find that the scaling parameter $\bm \gamma$ in the LayerNorm structure works as an outlier amplifier, which amplifies the outliers in the output. By extracting it, the activation becomes more robust for quantization. Then by further studying the clipping impact, we discover that the influence of final performance when clipping the outliers varies greatly, where some more aggressive outliers covering a large area can be clipped safely without accuracy degradation, but the accuracy can drop suddenly when the important outliers are clipped. More interestingly, though those less important outliers might present in a long tail form, they are only provided by a few tokens.

% In this paper, to suppress the outliers rather than walk around them, we make an in-depth analysis to investigate the inducement of the outliers and the impact of clipping the outliers.

% To be specific, we first exploit the inducement and find that the scaling parameter $\bm \gamma$ in the LayerNorm structure works as an outlier amplifier and strengthens the outliers in the output. By extracting it, the activation is more robust to quantization. Then we further study the impact of outlier clipping and discover that different outliers may have different impacts on the full-precision performance when they are clipped. More interestingly, the more aggressive outliers provided by a few tokens, such as the separator token, can be clipped sharply and safely without much accuracy degradation. 

Motivated by the analysis, we propose an outlier suppression framework to push the limit of low-bit Transformer language models. Such framework contains two key components: Gamma Migration and Token-Wise Clipping, corresponding to the above two findings. The Gamma Migration produces a more quantization-friendly model by migrating the outlier amplifier $\bm \gamma$ into subsequent modules in an equivalent transformation and bringing more robust activation for quantization without extra computation burden. The Token-Wise Clipping further efficiently finds a suitable clipping range with minimal final quantization loss in a coarse-to-fine procedure. The coarse-grained stage, which leverages the fact that those less important outliers only belong to a few tokens, can obtain a preliminary clipping range quickly in a token-wise manner. The fine-grained stage then optimizes it. Our proposed framework can be applied to different models and tasks, and coupled with existing methods. More essentially, the thought of outlier suppression shall shed new light on the study of NLP quantization.

% Motivated by these findings, we propose an outlier suppression framework to push the limit of low-bit Transformer language models, which suppresses the outliers by equivalently migrating the outlier amplifier and efficiently finding an appropriate clipping range. Such framework contains two key components: Gamma Migration and Token-Wise Clipping, corresponding to these two findings.Gamma Migration extracts the scaling parameter $\bm \gamma$ in LayerNorm and transfers it into subsequent modules with an equivalent transformation, significantly alleviating the outliers. 
% Thus we can quantize on a more robust activation with no extra computation overhead. 
% Then, Token-Wise Clipping further suppresses the unimportant outliers which still exist without the amplification. Compared to previous calibration methods, which ignore the outliers' importance or suffer from large time costs on those useless outliers, our method leverages the fact that those less important values only belong to a few tokens and first obtains a preliminary clipping range from a token perspective, then optimizes it in a fine-grained way. 
% Our proposed framework can be combined with existing methods, and the thought of outlier suppression shall shed new light on the study of NLP quantization.

To summarize, our contributions are as follows:

\begin{enumerate}[nosep, leftmargin=*]

    \item We delve into the inducement and clipping impact of outliers in the NLP models and draw two critical findings that help handle the bottleneck of transformer quantization.%One is that the scaling parameter of LayerNorm acts as an outlier amplifies that is bad for quantization. The other is that the more aggressive outliers provided by several tokens are not as important as their value suggests and can even be clipped without any accuracy drop in full-precision models.
    % \item We explore the inducement and importance of outliers in the NLP models and find that scaling parameter of LayerNorm leads to the widespread outliers and the importance of outliers is highly related with tokens.
    \item Based on the findings, an outlier suppression framework containing Gamma Migration and Token-Wise Clipping is proposed. This framework is efficient, easy to implement, and plug-and-play. 

    \item The Gamma Migration suppresses the outliers from the inducement aspect and produces a more quantization-friendly model without any extra inference time. It transfers the outlier amplifier in LayerNorm to the subsequent modules in an equivalent transformation and contributes to activation with less quantization error.
    \item The Token-Wise Clipping scheme suppresses the outliers from the aspect of importance and produces a superior clipping range efficiently. It can skip over those unimportant outliers quickly leveraging the large variance of token range and then focus on the influential area.
    % \item For the outliers induced by LayerNorm, we propose to suppress them with a Gamma Migration technique, which migrates the scaling parameter to the subsequent modules utilizing an equivalence transformation. Gamma Migration contributes to a quantization-friendly distribution and brings no extra inference time.
    % contributes to a quantization-friendly distribution and brings no extra computation overhead.
    %\item To detect an appropriate clipping value for the outliers, a Token-Wise Clipping scheme is designed which can skip over those unimportant outliers quickly from the token perspective and focus on the influential area with fine-grained learning. 
    % \item For the unimportant outliers, a Token-Wise Clipping scheme is designed to find the optimal clipping range from the token perspective. By first coarsely token-wise searching and then fine-grained learning, it can maintain the important outliers and eliminate the meaningless ones with minimal time cost.
    \item Extensive experiments on various NLP models (BERT, RoBERTa, BART) and tasks (text classification, question answering, and summarization) prove that our outlier suppression framework sets up a new state of the art for transformer quantization, and for the first time, pushes the 6-bit post-training quantization (PTQ) and 4-bit quantization-aware training (QAT) accuracy of BERT to the full-precision level.
    %加了一些关键词
    % PTQ: push the limit of 6-bit quantization without significant accuracy loss for BERT models. (GLUE benchmark, largest enhancement BERT: 29.31\%, RoBERTa: 25.62\%, BART: 30.73\%) 或者这边可以描述glue benchmark8个数据集average的提升。
    % QAT: push the limit of 4-bit quantization without significant accuracy loss without leveraging the distillation or data augmentation for BERT models. (largest enhancement 50.56
    % \%)
\end{enumerate}

\section{Preliminaries}
\label{sec_preliminary}
\textbf{Basic Notations. }We mark matrices as $\mX$ and vectors as $\vx$. Operator $\cdot$ denotes scalar multiplication, and $\odot$ is adopted for element-wise multiplication on matrices or vectors. Also, we use $\mW\vx$ as matrix-vector multiplication. Specifically, considering the tokens in NLP tasks, $\mX_{t, j}$ stands for the element at token $t$ and embedding $j$, and $\vx_t$ represents the embedding of token $t$.

\textbf{Quantizer. }Quantization usually includes two operations. 
% Here, we obey the quantizer definition proposed in \cite{Jacob_2018_CVPR}.
\begin{equation}
     \bar{x} = clip(\lfloor \frac{x}{s} \rceil + z, \,0,\,2^b-1),\ \ \
	 \hat{x} = (\bar{x} - z )\cdot s
\end{equation}
where $s$ (step size), $z$ (zero point) are quantization parameters, $b$ is the bit setting. The first operation called "Quant" maps continuous numbers ($x$) to discrete points ($\bar{x}$) for integer-arithmetic-only matrix computation. The second operation called "DeQuant" recovers it to $\hat{x}$ after multiplication.
\section{Outlier analysis}
\label{sec_observation}
For Transformer-based models, standard 6/8-bit PTQ or 4-bit QAT would cause severe accuracy degradation. Investigating each quantizer, we recognize that the output of LayerNorm structures and GELU functions hold some sharp outliers, which should be responsible for the large quantization error. Evidence and experimental results in \autoref{appendix_problematic_quantization_nodes}. 

%fix
To deeply investigate the relationship between the harmful outliers and quantization performance, we explore the underlying inducement and impact of clipping the outliers. Before that, some brief descriptions (see \autoref{appendix_outliers phenomenon} for detailed ones) about the outliers are given first to help understand the following two parts. The outliers show structured characteristics that they mainly gather at some certain embedding dimensions, and upon these dimensions, the outliers provided by unique tokens like the separate toke and comma even hold more aggressive values.

\begin{figure}[t]
    \leftskip1mm
    \begin{subfigure}[t]{0.19\textwidth}
        \centering
        \includegraphics[height=\textwidth]{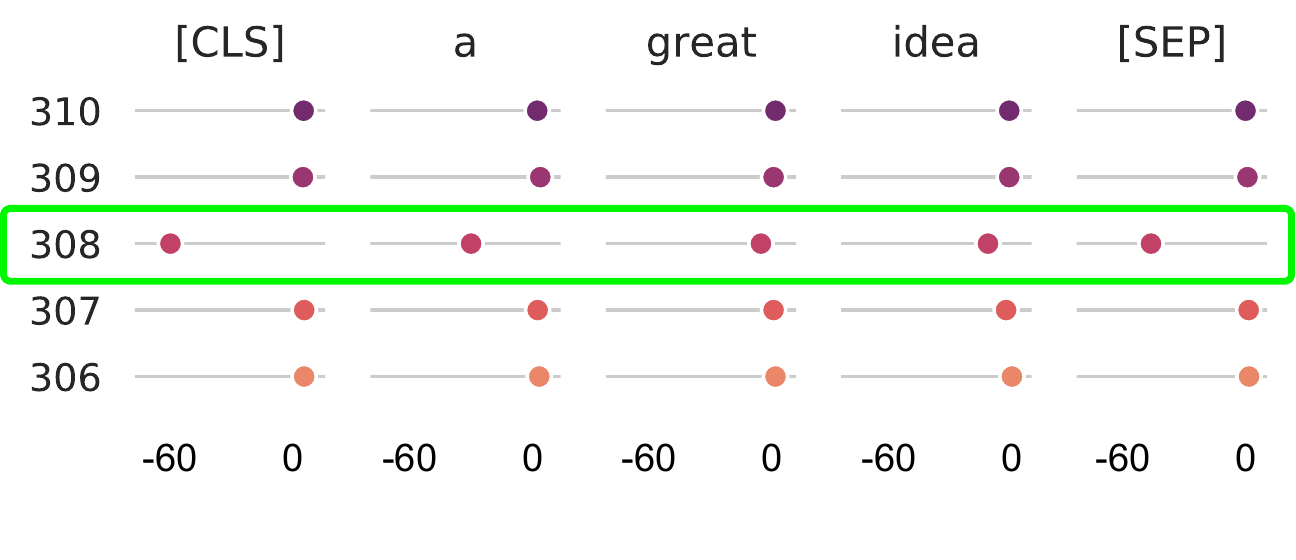}
        \caption{$\widetilde{\mX}$}
        \label{fig_motivation1_attn}
    \end{subfigure}
    \hspace{28mm}
    \begin{subfigure}[t]{0.19\textwidth}
        \centering
        \includegraphics[height=\textwidth]{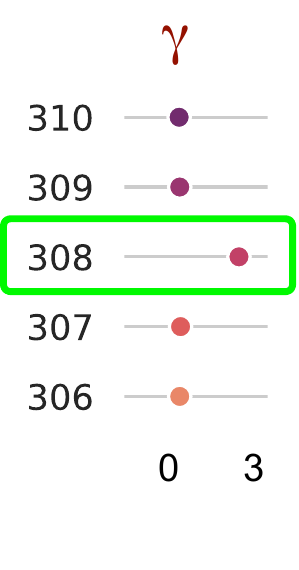}
        \caption{$\bm \gamma$}
        \label{fig_motivation1_gamma}
    \end{subfigure}
    \hspace{-7mm}
    \begin{subfigure}[t]{0.19\textwidth}
        \centering
        \includegraphics[height=\textwidth]{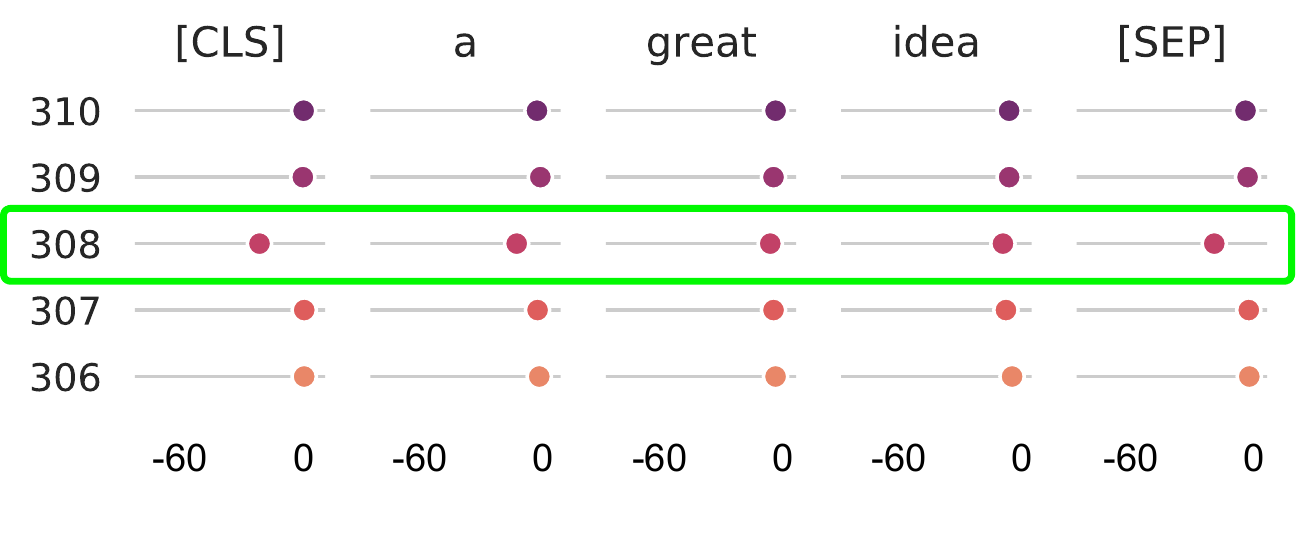}
        \caption{$\mX'$}
        \label{fig_motivation1_delay}
    \end{subfigure}
    % \hspace{300mm}
    \caption{Presentation of outliers over $\widetilde{\mX}$, $\bm \gamma$ and $\mX'$ of LayerNorm on BERT-SST-2. For example, at dimension 308, $\bm \gamma$ and $\widetilde{\mX}$ both have sharper values. By excluding $\bm \gamma$, it can be seen that $\mX'$ holds milder distribution than $\widetilde{\mX}$. More evidence is put in \autoref{appendix_outlier_inducement}. }
    \label{fig_motivation1}
\end{figure}

\subsection{Inducement of outliers}
\label{subsec_observation1}
% fix
For the inducement of outliers, we find that the scaling parameter in LayerNorm amplifies the outliers from embedding dimensions. And the phenomenon that some tokens have sharper outliers might be caused by the uneven token frequency in the pre-training phase (see \autoref{appendix_analysis_outlier}). In this part, we mainly explain the first inducement to solve these outliers from the origin. For another one, due to the high cost of adjusting the pre-training, we discuss the clipping impact in the next part to suppress these outliers from the clipping perspective.

Considering the challenges of quantizing the LayerNorm, the natural action is to dive into its internal structure. For token $t$ at $j^{th}$ embedding dimension, it first normalizes the input using mean ($\vu_t$) and variance ($\bm \sigma_t^2$) each forward pass, then scales and shifts the value with parameter $\bm \gamma_j$ and $\bm \beta_j$.

\begin{equation}
    \label{equation_ln}
    \mathbf{LayerNorm:} \ \ \ \ \widetilde{\mX}_{t,j} = \frac{\mX_{t,j}-\vu_t}{\sqrt{\bm\sigma_t^2 + \epsilon}} \cdot \bm \gamma_j + \bm \beta_j
\end{equation}

%amplification
Then, by observing the parameter distribution of LayerNorm, we surprisingly find that the multiplier $\bm \gamma$ (\autoref{fig_motivation1_gamma}) and the output $\widetilde{\mX}$ (\autoref{fig_motivation1_attn}) hold outliers at the same embedding dimensions. Besides, the adder $\bm \beta$ denotes a smaller range (e.g., (0,3)) compared to the output range (e.g., (-60, 0)), so we ignore it for identifying the key point. That is to say, $\bm \gamma$ plays a crucial part for the outliers in \autoref{fig_motivation1_attn}, especially can amplify the outliers across tokens by serving as a shared parameter.

%amplification move better
This observation enlightens us to remove the amplification effect by extracting $\bm \gamma$  from \autoref{equation_ln} and use the Non-scaling LayerNorm \autoref{equation_delay_ln}.
\begin{align}
    \label{equation_delay_ln}
   \mathbf{Non\operatorname{-}scaling\ LayerNorm:} \ \ \ \ \mX'_{t,j} & = \frac{\mX_{t,j}-\vu_t}{\sqrt{\bm\sigma_t^2 + \epsilon}} + \frac{\bm \beta_j}{\bm \gamma_j}
%   \widetilde{\mX}_{i,j} & = \mX'_{i,j} \cdot \bm \gamma_j
 \end{align}
\autoref{fig_motivation1_delay} and \autoref{fig_motivation1_attn} show that the output of the Non-scaling LayerNorm denotes a milder distribution with weaker outliers than the normal one.
It not only coincides with that $\bm \gamma$ does strengthen the outliers but also reveals that $\mX'$ behaves more friendly than $\widetilde{\mX}$ for quantization.
% Comparing \autoref{fig_motivation1_attn} and \autoref{fig_motivation1_delay}, it is obvious that $\mX'$ denotes a milder distribution with weaker outliers than $\mX$, which reveals that parameter $\bm \gamma$ does strengthen the outliers.

% quantitatively illustration
To quantitatively validate the more quantization-friendly distribution $\mX'$ holds, we adopt the cosine similarity metric to evaluate the quantization loss. From \autoref{tab_cosine_similarity}, the second row with higher similarity, namely less quantization error, explains that the quantization performance can be improved using Non-scaling LayerNorm.
% And we conjecture that the tensor $\mX'$ behaves more friendly to quantization than the complete output $(\widetilde{\mX}$). To quantitatively measure the quantization robustness, the cosine similarity, which evaluates the error that the quantizer injects to the real signal, is applied to the tensors before and after scaling transformation, seperately. It can be seen from \autoref{tab_cosine_similarity} that activations without $\bm \gamma$ multiplication actually induce much less quantization error than the whole one. These intuitive and numerical analysis encourage us that the quantization performance can be improved by pushing forward the quantizer.
\begin{table*}[htp!]
\footnotesize
    \centering
    \begin{adjustbox}{max width=\textwidth}
    \begin{tabular}{lcccccccccccc}
        \toprule
          \bf Tensor & \textbf{0} & \textbf{1} & \textbf{2} & \textbf{3} & \textbf{4} & \textbf{5} & \textbf{6} & \textbf{7} & \textbf{8} & \textbf{9} & \textbf{10} & \textbf{11} \\
        \midrule
            $\widetilde{\mX}$ & 97.16 & 97.03 & 97.61 & 94.37 & 93.41 & 93.53 & 93.31 & 93.61 & 94.56 & 95.62 & 96.13& 98.57 \\
        \midrule
            $\mX'$ & \textbf{99.23} & \textbf{99.22} & \textbf{99.11} & \textbf{99.02} & \textbf{98.99} & \textbf{99.00} & \textbf{98.99} & \textbf{98.83} & \textbf{98.70} & \textbf{99.05} & \textbf{99.44} & \textbf{99.07} \\
        \bottomrule
    \end{tabular}
    \end{adjustbox}
    \caption{Cosine similarity (\%) of the quantized value (6-bit) and the real signal for $\widetilde{\mX}$ and $\mX'$ across 12 LayerNorm after Multi-Head Attention on BERT-SST-2. Higher is better. More evidence in \autoref{appendix_outlier_inducement}. }
    \label{tab_cosine_similarity}
    % \vspace{-8pt}
\end{table*}

\subsection{Impact of outlier clipping}
\label{subsec_observation2}
% the importance of outlier are highly varied. The outliers which can be safely clipped are denoted by a few tokens. This can accelerate the ratio finding speed because those safely clipped outliers might occupy in a long tail form.

In this part, we explore the impact of clipping the outliers to design a method that can find an appropriate clipping range for quantization. The experiments are designed for the clipping impact on the accuracy and token of FP models.

\textbf{Impact on accuracy. }When clipping the outliers and evaluating the final performance, we find that the importance of outliers is highly varied. Take the outliers after GELU as an example here (others in \autoref{appendix_clipping_value}), \autoref{fig_clip_accuracy} shows that clipping the more aggressive outliers sharply (clipping signals in 10-100 to 10) even does not hurt the full-precision performance with accuracy still at 91.02, while the accuracy drops suddenly to 85.93 with too many outliers cut.

\begin{figure}[b]
    \centering
    \includegraphics[width=\textwidth]{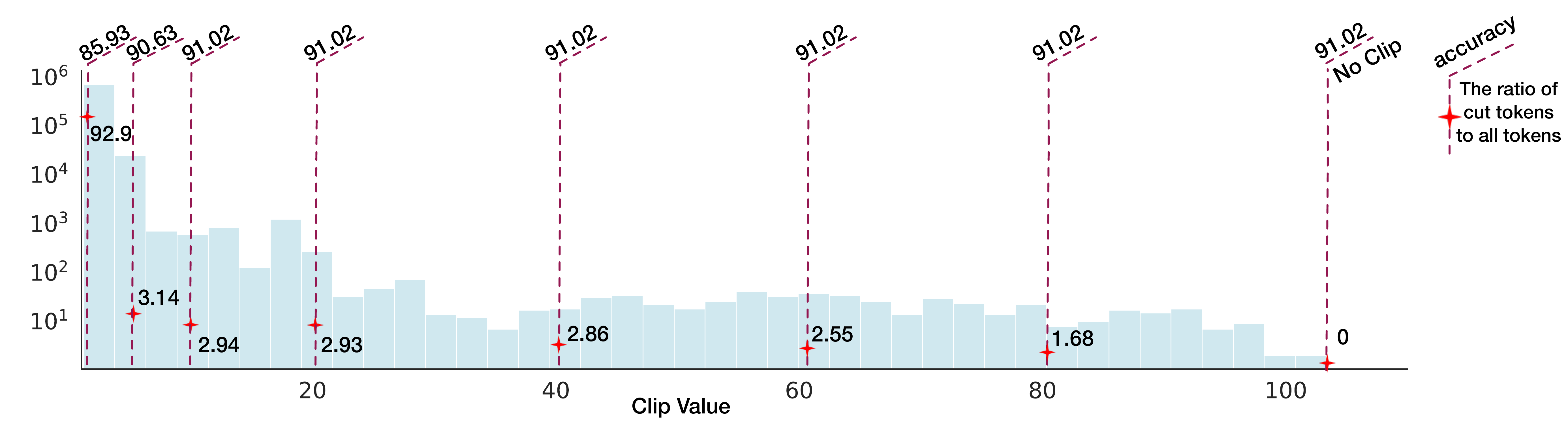}
     \caption{To detect the impact of clipping the outliers, we first draw the distribution using (mean + 3 * std) as its left border, then enumerate the value to cut the tensor on RoBERTa-QNLI. Red points reflect the proportion of clipped tokens. More evidence in \autoref{appendix_clipping_value}.} 
    \label{fig_clip_accuracy}
\end{figure}

\textbf{Impact on token. }Another key point is the unimportant outliers which can be clipped without even any accuracy drop in FP models only correspond to a few tokens. Motivated by \cite{bondarenko2021understanding}, they refer that the separator token [SEP] attends to larger values. We are also aware of the different ranges provided by different tokens. From the red points in \autoref{fig_clip_accuracy}, which represents the proportion of clipped tokens, it can be clearly seen that the more aggressive outliers though occupy a large range from 10 to 100 only matches with 3\% tokens. Destroying those sharper outliers belonging to a few tokens will not affect the performance.

The former investigation of accuracy impact suggests us taking the final performance into account to find a superior clipping range, where some local optimization methods like \cite{choukroun2019low} are not suitable here. The latter finding in token impact encourages us to leverage the token's indication to quickly skip over the unimportant area, especially when it presents in a long tail form where some methods like \cite{mckinstry2019discovering} suffer low efficiency. Based on these, we will introduce our method in \autoref{subsection_token_wise_clipping}.

\section{Method}
% Gamma Migration 
In this section, we propose our outlier suppression framework based on the above analysis. Firstly, the Gamma Migration technique is adopted to obtain a more quantization-friendly model by migrating the gamma into subsequent modules. Secondly, the Token-Wise Clipping further finds a suitable clipping range efficiently by leveraging the large variance of the token range.

\subsection{Gamma Migration}
\label{subsection_gamma_migration}
\begin{figure}[ht]
    \centering
    \includegraphics[width=\textwidth]{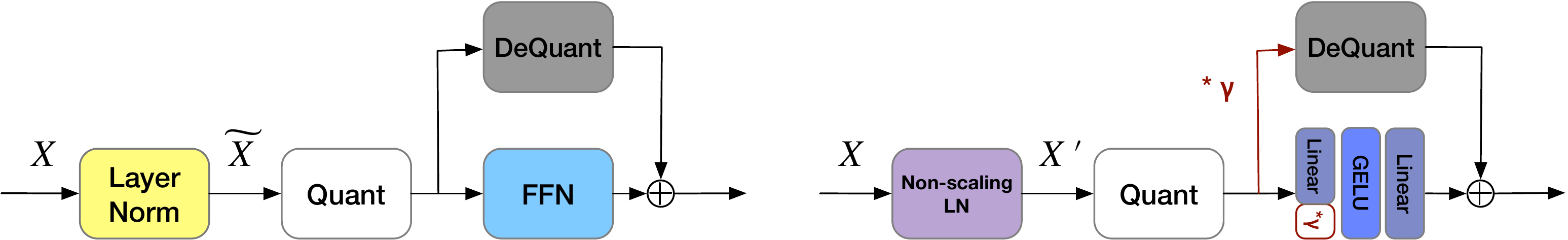}
    \caption{Comparison of the quantization flow before (left) and after (right) Gamma Migration. The original LayerNorm = the Non-scaling LayerNorm * $\bm \gamma$. For other detailed applications such as LayerNorm in encoder-decoder structure, see \autoref{fig_method_lnsplit_ffn}, \autoref{fig_method_lnsplit_encoder_decoder}.} 
    \label{fig_method_lnsplit}
\end{figure}

As pointed out in \autoref{subsec_observation1}, activation without going through the scaling parameter provides less quantization error. 
In this way, we split the LayerNorm function, migrate $\bm \gamma$ into follow-up structures and quantize the output of the Non-scaling LayerNorm. The transformation is equivalent for the FP model and brings more robust activation for the low-bit one. The overall flow is illustrated in \autoref{fig_method_lnsplit}.

\textbf{Migration equivalence on FP model. }Naturally, as referred in \autoref{equation_delay_ln}, we extract the parameter $\bm \gamma$ and transform the LayerNorm into Non-scaling one, thus seperate $\mX'_{t, j}$ from $\widetilde{\mX}_{t, j}$.
\begin{equation}
   \widetilde{\mX}_{t,j}  = \mX'_{t,j} \cdot \bm \gamma_j
\end{equation}
Since the residual connection is frequently adopted after LayerNorm (\cite{dai2019transformer,dehghani2018universal,clark2020electra}), it is necessary to illustrate the way to migrate parameter $\bm \gamma$ into two branches. To be specific, considering the LayerNorm after Multi-Head Attention (\autoref{fig_method_lnsplit}),  $\bm \gamma$ will be excluded from the LayerNorm and moved to the shortcut branch and weight of the next layer. Then the LayerNorm becomes the Non-scaling one, the shortcut branch establishes a new parameter $\bm \gamma$, and the weight of the next layer can absorb the $\bm \gamma$.
% To put it in practical terms, such as LayerNorm after Multi-Head Attention (\autoref{fig_method_lnsplit}), $\bm \gamma$ in LayerNorm is canceled out but reestablished on the shortcut branch and can be absorbed in the weight of next Layer.

Now, we show how the weight absorbs $\bm \gamma$. For linear layers, we have the following equation:
\begin{equation}
\label{equation_linear_transformation}
\mW(\vx\odot \begin{bmatrix}
\bm \gamma_1\\
\bm \gamma_2\\
...\\
\bm \gamma_n \end{bmatrix})=(\mW \odot \begin{bmatrix}
\bm \gamma_1 & \bm \gamma_2 & ... & \bm \gamma_n \\
\bm \gamma_1 & \bm \gamma_2 & ... & \bm \gamma_n \\
...\\
\bm \gamma_1 & \bm \gamma_2 & ... & \bm \gamma_n\end{bmatrix})\vx, 
\end{equation}
where $\vx$ serves as a column vector and $\bm \gamma \in \mathbb{R}^n$. The proof is available in \autoref{appendix_proof}. 
Because $\bm \gamma$ is a shared parameter, each token's embedding satisfies \autoref{equation_linear_transformation}, which promises success of transferring the $\bm \gamma$ into the next layer's weight.

\textbf{Quantization after migration. }Deriving from the above equivalent transformation, we outline
the quantization pattern after the migration process. From \autoref{fig_method_lnsplit}, the "Quant" process is employed at $\mX'$, then the quantized output engages in the matrix multiplication on one branch, multiplies parameter $\bm \gamma$ and experiences the "DeQuant" process on another branch. In fact, this means delaying the $\bm \gamma$ calculation from LayerNorm to the shortcut branch. Hence, this new design will not increase the computation overhead.
% In fact, considering the original fashion, which first multiplies $\bm \gamma$ and divides step size, and then experiences the "DeQuant" function, our method matches with delaying the $\bm \gamma$ multiplication. Hence, this new design will not increase the computation overhead.
% due to the subsequent "Add" operation.

% \textbf{Effect of migration. }We then analyze the effect of Gamma Migration on weight and activation respectively to that the activation quantization burden has been greatly alleviated with relatively a slight influence on weight. Theoretically, as presented in \autoref{fig_motivation1}, outliers emerge at the same embedding dimensions on $\bm \gamma$, activation before ($\mX'$) and after ($\widetilde{\mX}$) scaling function. In the original structure, the absolute max range of output can be actually rewritten as $|max(\mX')| * |max(\bm \gamma)|$. However, the weight matrix does not have the same embedding outlier phenomenon as the activation. Thus, in our method, the newly quantized activation range becomes $|max(\mX')|$ and weight range will not be amplified $|max(\bm \gamma)|$ times. Experimentally, \autoref{tab_cosine_similarity} in \autoref{subsec_observation1} has validated the favor to activation. We also calculate the cosine similarity for the changed weight and observe that $\bm \gamma$ has little impact on weight (\autoref{tab_weight_cosine_similarity}). 
\textbf{Effect of migration. }We then analyze the effect of Gamma Migration on weight and activation, respectively, to reveal that the activation quantization burden has been greatly alleviated with relatively a slight influence on weight. To begin with, suppose that the absolute max range of output in the original LayerNorm is $|max(\mX')| * |max(\bm \gamma)|$ for the reason that outliers emerge at the same embedding dimensions among $\bm \gamma$, activation before~$\mX'$  and after~$\widetilde{\mX}$ scaling function. For activation, extracting the $\bm \gamma$ will reduce the activation range by $|max(\bm \gamma)| $ times. And the results in \autoref{tab_cosine_similarity} have already validated the profit the transformation brings to activation. For weight, the weight matrix does not have the same embedding outlier phenomenon as the activation. Therefore, the weight range will not be amplified $|max(\bm \gamma)|$  times after the migration. Experimentally, we also calculate the cosine similarity for the changed weight and observe that $\bm \gamma$ has little impact on weight (\autoref{tab_weight_cosine_similarity}). 
% By the way, quantization-aware training can enjoy the benefit of Gamma Migration (\autoref{fig_loss}).

\begin{table*}[ht]
\footnotesize
    \centering
    \begin{adjustbox}{max width=\textwidth}
    \begin{tabular}{lcccccccccccc}
        \toprule
          \bf Tensor & \textbf{0} & \textbf{1} & \textbf{2} & \textbf{3} & \textbf{4} & \textbf{5} & \textbf{6} & \textbf{7} & \textbf{8} & \textbf{9} & \textbf{10} & \textbf{11} \\
        \midrule
            original weight & 99.95 & 99.95 & 99.95 & 99.95 & 99.95 & 99.95 & 99.95 & 99.95 & 99.95 & 99.95 & 99.95& 99.95 \\
        \midrule
            changed weight & 99.95 & 99.95 & 99.95 & 99.90 & 99.90 & 99.92 & 99.94 & 99.95 & 99.95 & 99.95 & 99.91 & 99.94 \\
        \bottomrule
    \end{tabular}
    \end{adjustbox}
    \caption{Cosine similarity (\%) between the quantized value (6-bit) and the real signal for original weight and the changed weight across 12 Intermediate layers on BERT-SST-2. It can be seen that there is little disparity between the two rows, especially compared with \autoref{tab_cosine_similarity}. }
    \label{tab_weight_cosine_similarity}
\end{table*}

\subsection{Token-Wise Clipping}
\label{subsection_token_wise_clipping}

\begin{figure}[b]
    \centering
    \includegraphics[width=\textwidth]{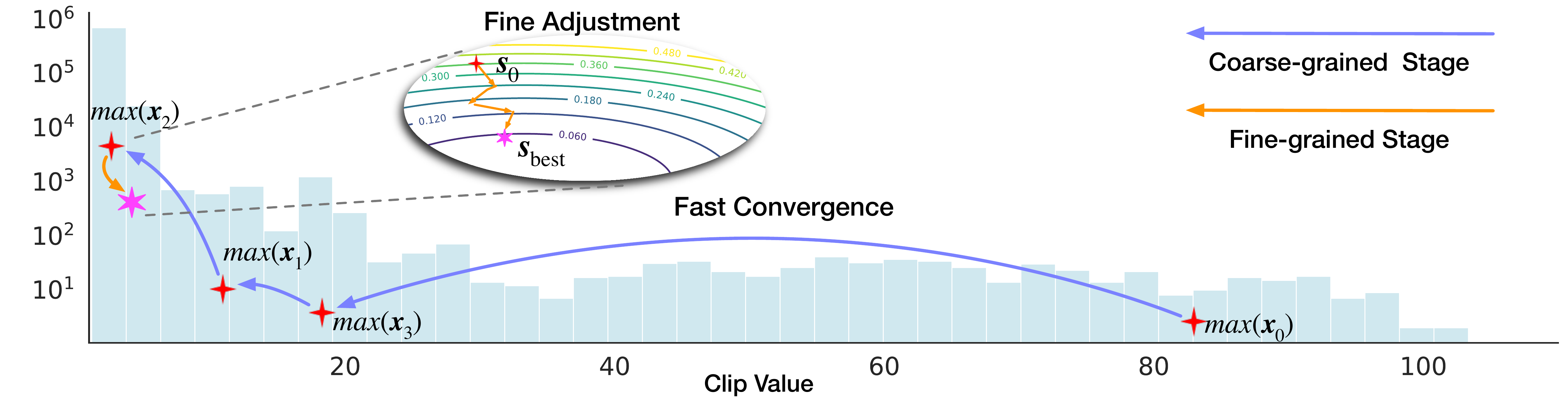}
    \caption{Flow diagram of the proposed Token-Wise Clipping} 
    \label{fig_method_twc}
\end{figure}
% final accuracy change, token information
Based on the analysis, we propose the Token-Wise Clipping method which considers the final loss when finding a clipping range and takes a coarse-to-fine paradigm to minimize it efficiently in a token-wise manner.

Regarding the very different accuracy impact of clipping the outliers, we search the clipping range, equivalently the step size $s$, which has the minimal distance between the final quantized output $\hat{f}(s)$ and the real one $f$ defined as \autoref{equation_quantization_loss}. To implement the process efficiently, especially when the unimportant outliers cover a wide area, a coarse-to-fine paradigm is designed below.

\begin{equation}
\label{equation_quantization_loss}
    L(s) = \|\hat{f}(s) -f\|_F^2,
\end{equation}

\textbf{Coarse-grained Stage. }At this stage, our aim is to quickly skip over the area where clipping causes little accuracy influence. According to \autoref{subsec_observation2}, the long tail area only matches with a few tokens. Therefore, we suggest using the max value of the embedding at token $t$ to be its representatives (min value as representatives for negative outliers). A new tensor with $T$ elements can be constructed by taking out the maximum signal for each token:
\begin{equation}
    \label{equation_collection_max}
    \vo^{u} = \{max(\vx_1),\, max(\vx_2),\, ...\,, \,max(\vx_{T}) \},
\end{equation}
where $\vo^u$ is marked as the collection of upper bounds, $\vo^l$ as the collection of lower bounds.

Then for a clipping ratio $\alpha$ on $\vo^u$, calculate the corresponding clipping value $c^u$ and use it to cut the tensor.
\begin{equation}
\label{equation_quantile}
    c^{u} = quantile (\vo^{u}, \alpha),
\end{equation}
where the quantile function computes the $\alpha\text{-}th$ quantiles of $\vo^{u}$.

Through grid search of token-wise clipping ratio, step size $s = \frac{c^{u} - c^{l}}{2^{b}-1}$ ($b$ is the bit-width) with minimal quantization loss \autoref{equation_quantization_loss} is obtained.
We mark it as $s_0$ for later optimization.

\textbf{Fine-grained Stage. }At this stage, our aim is to make some fine-grained adjustments in the critical area to further provide a guarantee for the final effect. In detail, with the initialization $s_0$, a learning procedure based on gradient descent is used to update parameter $s$ towards loss $L(s)$ with learning rate $\eta$, as described in \autoref{equation_gradient_descent}.

% In detail, a learning procedure is equipped to tune the step size $s$ with the initialization $s_0$ supplied by the coarse phase, where the  In detail, as defined in \autoref{equation_gradient_descent}, we use gradient descent to update parameter $s$ towards loss $L(s)$ with learning rate $\eta$.
% In detail, we optimize parameter $s$ towards reducing the distance between the final quantized output $\hat{f}(s)$ and the real one $f$ and use gradient descent with initialization $s_0$ and small learning rate $\eta$ to update $s$.
\begin{equation}
\label{equation_gradient_descent}
    s  = s - \eta \frac{\partial{L(s)}}{\partial{s}}
\end{equation}

% \textbf{Fine-grained Stage.} After coarse stage, learning procedure is equipped for fine-grained tuning. In detail, \autoref{equation_scale} uses the clipping range to calculate the step size, where $b$ is the bit setting. 

% Then towards reducing the final output distance between the quantized model and the real one, parameter $s$ is tuned by gradient descent.

\textbf{Benefits. }We mainly explain the benefits of the coarse-grained stage here from efficiency and quantization performance, where the experimental comparisons with other existing approaches are put in \autoref{appendix_twc}. For efficiency, because the wide range of outliers only corresponds to a few tokens, passing through the unimportant area from the token perspective needs much fewer iterations than from the value perspective. Moreover, the representative collection reduces the size of the tensor ($\vo^u$ distilled from $\mX$), so the method can run very fast each iteration. For quantization performance, the first coarse step has already produced a suitable clipping range (\autoref{ablation_study}), which offers a good initialization point for upcoming tuning.

\section{Experiments}
In this section, we conduct two sets of experiments to verify the effectiveness of our outlier suppression framework. \autoref{ablation_study} shows the effect of each component. \autoref{main_results} lists the results compared with other existing approaches across text classification, question answering, and summarization tasks. On the whole, we evaluate GLUE benchmark~\cite{wang2018glue}, SQuAD~\cite{rajpurkar2016squad,rajpurkar-etal-2018-know}, and XSum~\cite{narayan-etal-2018-dont} and CNN/DailyMail~\cite{nallapati-etal-2016-abstractive} across BERT, RoBERTa, and BART models. Here, 4-4-4 presents 4-bit weight, embedding, and activation. And the model size under a certain bit is put in \autoref{tab_model_size}.

\subsection{Experimental setup}
\textbf{Implementation details. } To begin with, we identify the quantization nodes and take a reasonable scheme like the one in FasterTransformer~\cite{FasterTransformer} (Details see \autoref{appendix_quantization_nodes}). For PTQ, equipping our framework, we use 256 samples to calibrate the model. For QAT, our methods work on the calibration phase and later are combined with LSQ+ \cite{bhalgat2020lsq+}, a strong baseline for the training phase. For training, hyper-parameters like learning rate are searched both for our methods and baseline techniques for fair comparisons. Details see \autoref{appendix_implementation_details}. 

\textbf{Baseline. } For PTQ, we compare with the prevalent calibration mechanisms including MinMax~\cite{google-whitepaper}, OMSE~\cite{choukroun2019low}, Percentile ~\cite{mckinstry2019discovering}, EasyQuant~\cite{wu2020easyquant} and PEG~\cite{bondarenko2021understanding}. For QAT, we present the results of Q-BERT ~\cite{shen2020q}, Q8BERT~\cite{zafrir2019q8bert} and PEG~\cite{bondarenko2021understanding}. Also, because our framework applying in QAT is coupled with LSQ+~\cite{bhalgat2020lsq+}, we show the results of the pure LSQ+, and another canonical quantization approach PACT~\cite{choi2018pact}. Last but not least, the results combined with knowledge distillation (KD) proposed in TernaryBERT~\cite{zhang2020ternarybert} are included as well.

 \begin{wrapfigure}{r}{3.8cm}
 \vspace{-1.8em}
    \includegraphics[width=\textwidth]{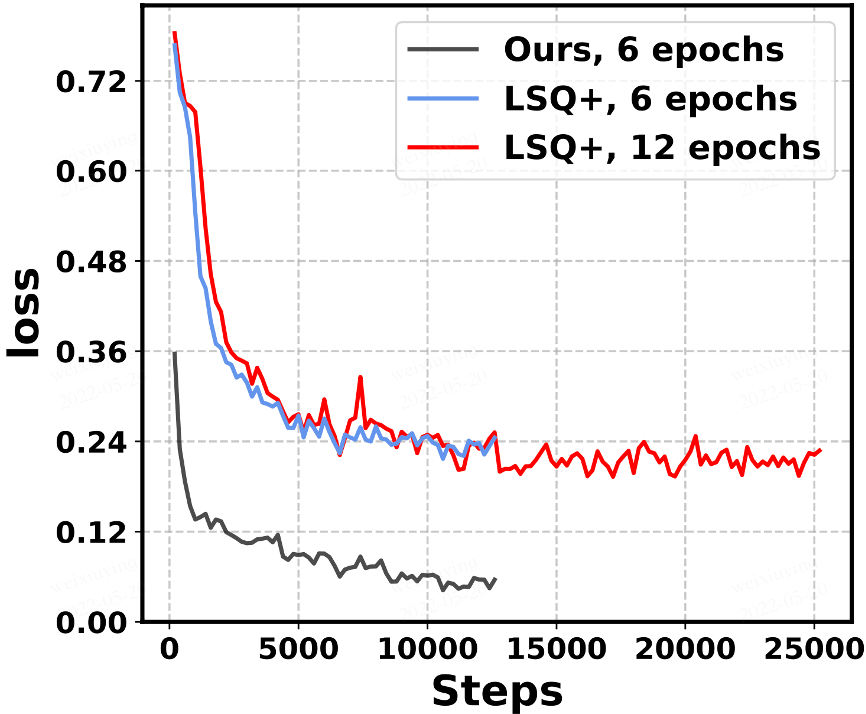}
      \caption{QAT fine-tuning process on BERT-SST-2.} 
     \label{fig_loss}
  \end{wrapfigure}

\subsection{Ablation Study}
\label{ablation_study}
In this subsection, we ablate the 
design elements in the proposed framework (\autoref{tab_ablation_study}).
As a general plug-in module, Gamma Migration helps both the MinMax and Token-Wise Clipping. And the Token-Wise Clipping also surpasses the baseline by a large margin: 17.53\% on QNLI, 13.22\% on MRPC (comparisons with other calibration algorithms see \autoref{appendix_twc}). About the phenomenon that the fine-grained stage sometimes does not improve much upon the coarse-grained one, we think it's due to the already good enough results produced by the coarse step.

Besides, \autoref{fig_loss} conveys that with a good initialization point provided by our framework, the training of QAT becomes much faster and easier.

\begin{table*}[ht]
    \centering
    
    \begin{adjustbox}{max width=\textwidth}
    \renewcommand{\arraystretch}{1.2}
    \begin{tabular}{lccccccccc}
        \hline
          \multirow{2}{*}{\bf Method} & \textbf{CoLA} & \textbf{MNLI} & \textbf{MRPC} & \textbf{QNLI} & \textbf{QQP} & \textbf{RTE} & \textbf{SST-2} & \textbf{STS-B} \\
        & (Matt.) & (acc m/mm) & (f1/acc) & (acc) & (f1/acc) & (acc) & (acc) & (Pear./Spear.) \\
        \toprule
         FP32                      & 62.50 & 87.75/87.23 & 93.1/90.44 & 92.68 & 88.78/91.6 & 80.51 & 95.18 & 91.04/90.72 \\
        \hline
         Baseline (MinMax)  & 0.0 & 34.9/35.0 &  71.64/67.4 & 62.13 &  51.88/74.37 & 49.82 & 77.87 & 44.11/46.74 \\ % 392.35
         MinMax + Gamma Migration     & 0.0 & 53.53/54.64 & \bf 87.97/82.84 & 78.56 & 78.04/85.3 & 55.6 & 85.67 &  61.03/63.22\\ % 500.4
         Token-Wise Clipping (Coarse)  & 34.95 & 80.56/80.84 & 85.05/79.41 & 79.46 &  85.96/89.31 & 66.43 &  91.63 & 82.03/82.45\\
         Token-Wise Clipping   & 37.64  & 81.13/81.26 & 85.59/79.9 & 79.66 & 85.83/89.26  & 64.62 &  91.63 & 83.10/83.51\\
         Gamma Migration + Token-Wise Clipping  & \bf 46.35 & \bf 83.38/83.32 &  87.50/83.33 & \bf 86.82 & \bf 86.82/90.01 & \bf 67.51 & \bf 92.2 & \bf86.83/86.93\\
        \bottomrule
        % \hline
    \end{tabular}
    \end{adjustbox}
    % }
    \caption{Results of our proposed Gamma Migration and Token-Wise Clipping for RoBERTa with 6-bit PTQ. }
    \label{tab_ablation_study}
    % \vspace{-8pt}
\end{table*}

\subsection{Main Results}
\label{main_results}
\subsubsection{Results on classification tasks}
\begin{table*}[ht]
    \centering
    
    \begin{adjustbox}{max width=\textwidth}
    \renewcommand{\arraystretch}{1.2}
    \begin{tabular}{lcccccccccc}
        \hline
          \multirow{2}{*}{\bf Method} & {\bf Bits} & \textbf{CoLA} & \textbf{MNLI} & \textbf{MRPC} & \textbf{QNLI} & \textbf{QQP} & \textbf{RTE} & \textbf{SST-2} & \textbf{STS-B} & \multirow{2}{*}{\bf Avg.} \\
        & (W-E-A) & (Matt.) & (acc m/mm) & (f1/acc) & (acc) & (f1/acc) & (acc) & (acc) & (Pear./Spear.) & \\
        \hline
        % 8-8-8 & \cite{bondarenko2021understanding} & & & & & &\\
        % \hline

        BERT                                            & 32-32-32 & 59.60 & 84.94/84.76 & 91.35/87.75 & 91.84 & 87.82/90.91 & 72.56 & 93.35 & 89.70/89.28 & 83.83 \\
        \hline
        \hspace{0.5em} MinMax & 8-8-8 & 57.08 & 82.77/83.47 & 89.90/85.78 & 90.76 & 87.84/90.74 & 69.68 & 92.78 & 86.83/88.56 & 82.28 \\ 
         \hspace{0.5em} OMSE \cite{choukroun2019low}    & 8-8-8 & 57.15 & 84.04/84.29 & 90.10/85.78 & 91.12 & 87.64/90.54 & 72.20 & 93.23 & 87.90/88.65 & 82.90 \\ 
        %  \hspace{0.5em} PEG \cite{bondarenko2021understanding} & 8-8-8 & 58.58 & 84.88/84.79 & 91.51/87.99 & 91.62 & 87.79/90.88 & 72.56 & 93.12 & 86.80/88.27 \\
         \hspace{0.5em} \textbf{Ours}  & 8-8-8 & \textbf{61.64} & \textbf{84.38/84.53} & \textbf{91.44/87.75} & \textbf{91.49} & \textbf{87.92/90.77} & \textbf{72.20} & \textbf{93.81} & \textbf{89.23/89.01} & \textbf{83.96} \\
         \hspace{0.5em} OMSE  & 6-6-6 & 35.44 & 74.00/73.30 & 81.54/76.47 & 84.66 & 76.07/82.12 & 64.26 & 86.27 & 85.57/86.05 & 73.52 \\ 
         \hspace{0.5 em} Percentile \cite{mckinstry2019discovering} & 6-6-6 & 37.32 & 72.40/71.69 & 85.09/79.90 & 79.37 & 72.58/80.19 & 61.73 & 87.27 & 86.38/87.29 & 72.93 \\
         \hspace{0.5 em} EasyQuant \cite{wu2020easyquant} & 6-6-6 & 38.16 & 75.82/75.66 & 82.51/77.45 & 84.94 & 75.31/81.81 & 65.34 & 87.27 & 85.50/86.33 & 74.49  \\
        %  \hspace{0.5em} PEG  & 6-6-6 & 55.34 & 82.00/82.15 & 86.49/81.62 & 87.79 & 85.10/88.33 & 70.40 & 90.83 & 87.58/87.93 \\
         \hspace{0.5em} \textbf{Ours}            & 6-6-6 & \textbf{54.40}  &  \textbf{82.02/81.69} & \textbf{87.45/83.33} & \textbf{89.82} & \textbf{84.69/88.94} & \textbf{70.76} &\textbf{91.86}  & \textbf{88.65/88.55} & \textbf{81.19} \\
         \midrule
         \hspace{0.5em} PEG \cite{bondarenko2021understanding} $^\clubsuit$ & 8-8-8 & 59.43 & 81.25 & 88.53 & \bf{91.07} & \bf 89.42 & 69.31 & 92.66 & 87.92 & 82.45 \\
         \hspace{0.5em} \textbf{Ours} $^\clubsuit$ & 8-8-8 & \bf{59.83} & \bf 82.93/82.59 & \bf \bf 91.33/87.99 & 90.02 & 87.45/90.34 & \bf 70.04 & \bf 92.66 & \bf 88.42/88.81 & \bf 82.81 \\

         \hspace{0.5em} PEG $^\clubsuit$ & 6-6-6 & 9.46 & 32.44/32.77 & 83.64/78.43 & 49.46 & 29.93/62.97 & \bf 70.76 & 90.14 & 52.79/53.22 & 54.11 \\
         \hspace{0.5em} \textbf{Ours} $^\clubsuit$ & 6-6-6 & \bf 42.27 & \bf 78.54/78.32 & \bf 85.33/81.13 & \bf 85.36 & \bf 78.47/84.66 & 68.59 & \bf 91.74 & \bf 87.33/87.19 & \bf 77.31 \\
         \hline
         RoBERTa                       & 32-32-32 & 62.50 & 87.75/87.23 & 93.1/90.44 & 92.68 & 88.78/91.6 & 80.51 & 95.18 & 91.04/90.72 & 86.40 \\
        \hline
         \hspace{0.5em} MinMax & 8-8-8 & 41.62 & 87.52/86.88 & 91.56/88.48 & 92.11 & 88.60/91.44 & 76.90 & 94.82 & 91.00/90.66 & 82.94 \\ 
         \hspace{0.5em}   OMSE           & 8-8-8 & 38.59 & 87.32/87.14 & 92.39/89.46 & 92.51 & 87.95/90.95 & 76.53 & 94.61 & 90.95/90.65 & 82.58 \\ 
         \hspace{0.5em}   \textbf{Ours}            & 8-8-8 & \bf 62.50 & \bf 87.61/87.31 & \bf 92.39/89.46 & \bf 92.53 & \bf 88.64/91.49 & \bf 78.34 & \bf 94.95 & \bf 91.08/90.73 & \bf 85.96 \\
         \hspace{0.5em}   OMSE           & 6-6-6 &  1.81 & 72.89/72.65 & 85.38/78.68 & 76.53 & 85.24/88.94 & 64.26 &91.17  & 80.81/81.99 & 69.63 \\ 
         \hspace{0.5 em} Percentile   & 6-6-6 & 20.73 & 72.23/73.68 & 84.83/78.43 & 77.16 & 82.21/87.44 & 62.82 & 88.19 & 79.41/79.64 & 70.98 \\
         \hspace{0.5 em} EasyQuant  & 6-6-6 & 9.28 & 74.96/75.87 & 84.31/76.47 & 74.04 & 85.52/89.12 & 62.45 & 89.56 & 80.89/82.38 & 70.01 \\
         \hspace{0.5em } \bf Ours & 6-6-6 & \bf 46.35 & \bf 83.38/83.32 & \bf 87.50/83.33 & \bf 86.82 & \bf 86.82/90.01 & \bf 67.51 & \bf 92.2 & \bf86.83/86.93 & \bf 79.62 \\
        %  \hspace{0.5em}   Ours            & 6-6-6 &  \textbf{46.35} & \textbf{83.49/83.67} & \textbf{87.30/82.60}  & \textbf{85.85} & \textbf{86.69/89.91} & \textbf{64.26}  &\textbf{92.09}  &  \textbf{87.21/87.27}\\
        \hline
        BART                         & 32-32-32 & 56.32 & 86.45/86.55 & 91.37/87.50 & 92.31 & 88.34/91.39 & 79.06 & 93.35 & 90.11/89.94 & 84.61 \\ 
        %  \cline{2-10}
        \hline
        \hspace{0.5em} MinMax & 8-8-8 & 55.38 & 85.87/86.14 & 89.44/85.29 & 91.20 & 88.07/91.24 & 77.98 & 93.69 & 89.90/89.73 & 83.89 \\ 
         \hspace{0.5em} OMSE           & 8-8-8 & 54.56 & 85.6/86.25 & 90.31/86.27 & 90.74 & 88.21/91.3 & 78.7 & 93.58 & 90.07/89.88 & 83.94 \\ 
         \hspace{0.5em} \bf Ours            & 8-8-8 & \textbf{55.53}  & \textbf{86.28/86.17} & \textbf{90.40/86.52} & 	\textbf{91.47}	 & \textbf{88.25/91.35} & \textbf{80.51} & \textbf{93.92} & \textbf{90.20/89.95} & \bf 84.50 \\
         \hspace{0.5em} OMSE           & 6-6-6 & 31.06 & 41.92/42.08 & 56.37/54.36 & 52.72 & 78.96/86.02 & 51.99 & 87.39 & 84.38/85.69 & 61.01 \\ 
         \hspace{0.5 em} Percentile  & 6-6-6 & 26.21 & 74.72/75.29 & 83.52/74.26 & 53.71 & 82.64/87.48 & 67.15 & 87.96 & 63.99/65.01 & 67.31 \\
         \hspace{0.5 em} EasyQuant  & 6-6-6 & 25.66 & 43.48/43.27 & 59.26/59.56 & 50.76 & 81.89/87.67 & 52.71 & 87.73 & 85.39/86.74 & 61.31 \\
         \hspace{0.5em} \bf Ours            & 6-6-6 & \textbf{44.51} & 	\textbf{82.46/82.98}	 & \textbf{86.41/80.88} & \textbf{86.34} & \textbf{83.60/88.45} & \textbf{71.12} & \textbf{90.94} & \textbf{87.56/87.38} & \bf 79.10 \\
        \bottomrule
        
        % \hline
    \end{tabular}
    \end{adjustbox}
    % }
    \caption{PTQ performance on GLUE benchmark. $^\clubsuit$: results taking the same quantization nodes with PEG ~\cite{bondarenko2021understanding} for fair comparisons. For the percentile, we search the hyper-parameter in [0.999, 0.9999, 0.99999] and report the best on dev set. }
    \label{tab:ptq_glue_results}
\end{table*}
\textbf{PTQ. } 
% We choose BERT, RoBERTa and BART models to validate the performance on classification tasks and summarize the results in \autoref{tab:ptq_glue_results}. 
\autoref{tab:ptq_glue_results} shows the results of PTQ on GLUE tasks.
For 8-bit BERT models, although previous methods generally behave well, our methods can still achieve satisfying outcomes even on small datasets such as CoLA (4.49\% upswings) and STS-B (1.33\% upswings). To fully exploit the limit, we try a more inspiring setting with weight and activation quantized to 6-bit. It can be seen that ours is indeed close to FP value within 2.64\% overall. Meanwhile, we also compare with PEG
\cite{bondarenko2021understanding} fairly by taking their quantization nodes. To be noted, their per-embedding-group (PEG) quantization certainly brings extra computation overhead and might not be available on real deployment while ours brings favorable results and can enjoy lossless acceleration on hardware. Besides, the experimental results on RoBERTa and BART consistently demonstrate our superiority whereas existing methods suffer from a non-negligible accuracy drop. On average, ours achieves up to 8.64\% and 11.79\% better accuracy on RoBERT and BART. To conclude, our proposed methods push the limit of 6-bit quantization to a new state of the art. 

\textbf{QAT. }In particular, we prove the compatibility of our methods on QAT. \autoref{tab:qat_glue_results} lists the results on BERT, other see \autoref{appendix_qat}. In a much harder setting  (4-4-4 bit quantization), our outlier suppression framework wins near-floating-point performance with a reduction of 2.70\% on average on 4-bit quantization. Yielding a good initialization, ours obtain an acceptable accuracy drop (0.7\% on QQP, 1.7\% on MNLI) without any distillation and data augmentation trick, versus 4.19\% and 3.16\% of LSQ+. Furthermore, ours still enables performance improvements working with knowledge distillation, especially at 2-bit weight and embedding. 
\begin{table*}[ht]
    \centering
    
    \begin{adjustbox}{max width=\textwidth}
    \renewcommand{\arraystretch}{1.2}
    \begin{tabular}{lcccccccccc}
        \toprule
          \multirow{2}{*}{\bf Method} & {\bf Bits}  & \textbf{CoLA} & \textbf{MNLI} & \textbf{MRPC} & \textbf{QNLI} & \textbf{QQP} & \textbf{RTE} & \textbf{SST-2} & \textbf{STS-B} & \multirow{2}{*}{\bf Avg.}\\
        & (W-E-A)  & (Matt.) & (acc m/mm) & (f1/acc) & (acc) & (f1/acc) & (acc) & (acc) & (Pear./Spear.) & \\
        \midrule
        % 8-8-8 & \cite{bondarenko2021understanding} & & & & & &\\
        % \hline

        BERT                                             & 32-32-32  & 59.60 & 84.94/84.76 & 91.35/87.75 & 91.84 & 87.82/90.91 & 72.56 & 93.35 & 89.70/89.28 & 83.83 \\
        \midrule
         Q8BERT \cite{zafrir2019q8bert} & 8-8-8 & 58.48 & - & 89.56/- & 90.62 & 87.96/- & 68.78 & 92.24 & 89.04/- & -\\
         Q-BERT \cite{shen2020q}                & 8-4-8 & - & 78.08/78.96 & - & - & - & - & 85.55  & - & - \\
         PACT \cite{choi2018pact}    & 4-4-8  & 55.23 & 83.98/83.90 & 91.58/88.24 & 91.12 & 88.19/91.20 & 71.84 & 91.86 & 89.73/89.27 & 82.89 \\
         LSQ+ \cite{bhalgat2020lsq+}    & 4-4-8 & 57.70 & 84.17/84.02 & 89.75/85.78 & 91.27 & 88.18/91.16 & 70.76 & 91.97 & \bf 89.74/89.3 & 82.84 \\
        %  \hspace{0.5em}  \cite{bondarenko2021understanding} $^\clubsuit$ & 4-4-8 & 57.22 & 83.69 & 87.77 & 91.29 & 89.64 & 70.04 & 92.32 & 89.13 \\
         PEG \cite{bondarenko2021understanding}  & 4-4-8 & 57.42 & 84.22/84.52 & 89.90/85.78 & 90.46 & 88.15/91.25 & 67.87 & 92.78 & 89.36/88.95 & 82.45
         \\
        %  \hspace{0.5em} GOBO \cite{zadeh2020gobo}$^\ast$ & 4-4-32 & - & 84.45 & - & - & - & - & - & 88.33 \\
        \bf Ours                            & 4-4-8 & \bf 61.06 & \bf 84.82/84.89 & \bf 91.26/87.75 & \bf 91.41 & \bf 88.45/91.40 & \bf 73.65 & \bf 92.55 & 89.71/89.24 & \bf 84.05\\
        %  \hspace{0.5em}  Ours $^\clubsuit$              & 4-4-8 & 59.57 & 85/84.31 & 87.75/91.07 & 91.31 & 88.35/91.32 & 72.2 & 92.43 & 89.57/89.2 \\
         
        %  PEG \cite{bondarenko2021understanding}  & 4-4-4 & 0.0 & 35.45/35.22 & 81.22/68.38 & 49.46 & 0.0/63.18 & 52.71 & 76.26 & nan/nan 
        %  \\
         PEG  & 4-4-4 & 0.0 & 35.45/35.22 & 81.22/68.38 & 49.46 & 0.0/63.18 & 52.71 & 76.26 & nan/nan & - 
         \\
        % \hspace{0.5em}  \cite{bondarenko2021understanding} $^\clubsuit$ & 4-4-4 & 0.0 & 35.45/35.22 & 0.0/31.62 & 49.46 & 0.0/63.18 & 52.71 & 49.08 & -0.0219/-0.0199 \\
         PACT     & 4-4-4 & 0.0 & 74.17/74.85 & 84.97/80.15 & 87.31 & 81.68/86.14 & 62.09 & 83.03 & 81.64/81.43 & 69.37 \\
         LSQ+                              & 4-4-4 & 0.0 & 81.40/81.97 & 88.34/83.82 & 88.10 & 83.11/87.24 & 64.62 & 82.34 & 84.16/83.75 & 71.49 \\ 
         \bf Ours  & 4-4-4 & \textbf{50.56} & \textbf{83.05/83.24} & \textbf{89.08/84.31} & \textbf{89.88} & \textbf{87.00/90.33} & \textbf{70.76} & \textbf{91.86} & \textbf{87.64/87.36} & \textbf{81.13} \\
        
         \midrule
          PEG $^\clubsuit$ $^\ast$ & 4-4-8 & 57.22 & 83.69 & 87.77 & 91.29 & 89.64 & 70.04 & 92.32 & 89.13 & 82.64 \\
         \bf Ours $^\clubsuit$              & 4-4-8 & \bf 59.57 & \bf 85.00/84.31 & \bf 91.07/87.75 & \bf 91.31 & \bf 88.35/91.32 & \bf 72.20 & \bf 92.43 & \bf 89.57/89.20 &\bf 83.60 \\
          PEG $^\clubsuit$ & 4-4-4 & 0.0 & 35.45/35.22 & 31.62/0.0 & 49.46 & 0.0/63.18 & 52.71 & 49.08 & -0.0219/-0.0199 & 29.25 \\
         \bf Ours $^\clubsuit$ & 4-4-4 & \textbf{51.93} & \textbf{83.03/83.24} & \textbf{89.39/85.05} & \textbf{90.33} & \textbf{87.38/90.62} & \textbf{72.56} & \textbf{91.74} & \textbf{88.36/87.91} & \textbf{81.76} \\
         
         \midrule
         LSQ+(+KD)          & 4-4-4 & 14.98 & 83.59/84.06 & \bf 92.47/89.46 & 91.16 & 87.96/91.01 & 67.87 & 85.55 & 84.17/83.96 & 75.99\\
         \bf Ours(+KD)       & 4-4-4 & \bf 56.67 & \bf 84.50/84.65 &  91.61/88.24 & \bf 91.45 & \bf 88.59/91.42 & \bf 74.37 & \bf 92.55 & \bf 89.13/88.78 &\bf 83.56\\
         LSQ+(+KD)     & 2-2-4 & 9.44 & 83.45/83.38 & 88.03/82.60 & 90.66 & 87.1/90.36 & 55.60 & 83.60 & 36.69/35.89 & 66.63 \\
         \bf Ours(+KD)      & 2-2-4 & \textbf{47.02} & \textbf{84.56/84.31} & \textbf{90.97/87.25} & \textbf{90.83} & \textbf{88.08/91.12} & \textbf{65.70} & \textbf{91.86} & \textbf{86.12/85.78} & \textbf{80.56} \\
         \bottomrule

    \end{tabular}
    \end{adjustbox}
    % }
    \caption{Comparison among different QAT strategies with low-bit activation on GLUE benchmark for BERT. 
    % For 4-bit activation, we adapt the knowledge distillation approach refered in \cite{zhang2020ternarybert} and compare with others. 
    $^\clubsuit$: results taking the same quantization nodes with PEG ~\cite{bondarenko2021understanding} for fair comparisons. $^\ast$: combined score for MNLI, MRPC, QQP and STS-B.}
    \label{tab:qat_glue_results}
    % \vspace{-8pt}
\end{table*}
\begin{table*}[b]
\footnotesize
    %\caption{Comparison among typical post-training quantization strategies in terms of mAP on MS COCO. Note that refer to \textsc{Brecq}, we didn't quantize head and keep the first and last layer in backbone to 8-bit. Other notations align the upper table. }
    \caption{Comparison among typical PTQ approaches in terms of f1/em on SQuAD. 
    % Other notations align \autoref{tab:ptq_glue_results}. 
    }
    \centering
    \begin{adjustbox}{max width=\textwidth}
    \begin{tabular}{lcccccccccc}
    \toprule
    \multirow{2}{6em}{\textbf{Method}} & \textbf{Bits} & \multicolumn{2}{c}{\textbf{BERT}} & \multicolumn{2}{c}{\textbf{RoBERTa}} & \multicolumn{2}{c}{\textbf{BART}}\\
    \cmidrule(l{2pt}r{2pt}){3-4}   
    \cmidrule(l{2pt}r{2pt}){5-6} 
    \cmidrule(l{2pt}r{2pt}){7-8} 
    & (W-E-A) & SQuAD v1.1 & SQuAD v2.0 & SQuAD v1.1 & SQuAD v2.0 & SQuAD v1.1 & SQuAD v2.0\\
    \midrule
    Full Prec. & 32-32-32 & 88.28/80.82 & 77.34/73.60 & 92.25/85.83 & 83.30/80.26 & 91.63/84.79 & 80.82/77.41 \\
    \midrule
    OMSE \cite{choukroun2019low} & 8-8-8 & \textbf{87.90/80.16} & 76.88/73.08 &  91.48/84.53 & 82.53/79.41 & 90.49/83.11 & 79.62/76.12 \\
    \bf Ours  & 8-8-8 & 87.60/79.80 & \textbf{76.93/73.14} & \textbf{91.57/84.86} & \textbf{82.94/79.72} & \textbf{91.08/84.07} & \textbf{80.55/77.04} \\
    \midrule
    OMSE  & 6-6-6  & 79.77/69.10 & 67.52/63.09 & 70.64/58.80 & 45.80/39.95 & 81.44/70.61 & 67.89/63.29 \\
    Percentile \cite{mckinstry2019discovering} & 6-6-6   & 78.55/67.14 & 69.12/65.64 & 67.24/53.28 &   56.38/51.58 & 82.45/72.87 & 68.44/63.29 \\
    EasyQuant \cite{wu2020easyquant} & 6-6-6  & 80.47/70.08 & 71.95/68.06 & 67.85/55.92 & 47.99/42.21 & 82.41/71.72 & 69.93/64.94 \\
    \bf Ours  & 6-6-6 &  \textbf{84.48/75.53} & \textbf{74.69/70.55} & \textbf{80.79/70.83}  & \textbf{68.47/64.10} & \textbf{83.68/75.34} & \textbf{74.44/70.36}\\
    \bottomrule
    \end{tabular}
    \end{adjustbox}
    \label{tab_ptq_squad}
    % \vspace{-8pt}
\end{table*}
\subsubsection{Results on question answering tasks}
To demonstrate the wider applicability of our methods, we evaluate them on SQuAD datasets. When going down to 6-bit quantization, the performance of other methods drastically drops. Ours still outperforms them by over 4.73\% and 15.55\% on BERT and RoBERTa on SQuAD v1.1. Also, the boost can be 12.31\% and 4.96\% on RoBERTa and BART on SQuAD v2.0.
\subsubsection{Results on summarization tasks}
% \input{tables/ptq_summ_results}

% \input{tables/ptq_squad_results}

% We present the PTQ results of BART on CNN DailyMail and XSum in \autoref{tab_ptq_summ}. The results show that our methods are also advanced on summarization tasks. 
It is of high value to validate the effect of our methods on summarization tasks. We choose classical datasets CNN/DailyMail and XSum and report the ROUGE 1/2/L score of BART. \autoref{tab_ptq_summ} illustrates that our approaches also benefit the encoder-decoder models, and can bring a near-floating-point performance on 8-bit and about 4\% enhancement on 6-bit.  \begin{table*}[t!]
      \begin{adjustbox}{valign=t,max width=\linewidth}
      \begin{tabular}{lcccccc}\toprule 
         Method & Bits(W-E-A) & CNN DailyMail & XSum  & Bits(W-E-A) & CNN DailyMail & XSum\\ 
         \midrule
         Full prec. & 32-32-32 & 45.62/22.85/42.88 & 42.82/20.11/34.99 & 32-32-32 & 45.62/22.85/42.88 & 42.82/20.11/34.99\\
         \midrule
         OMSE \cite{choukroun2019low} &8-8-8 & 44.89/22.03/42.18 & 41.58/18.77/33.73  &6-6-6 & 37.56/15.46/34.92 & 16.11/2.13/12.22 \\ 
         Percentile \cite{mckinstry2019discovering} &8-8-8 & 44.67/21.74/41.81 & 41.47/18.67/33.61  & 6-6-6 & 37.02/15.31/34.45 & 30.10/9.43/22.70 \\  
         EasyQuant \cite{wu2020easyquant} &8-8-8 & 44.98/22.07/42.24 & 41.65/18.81/33.77  & 6-6-6 & 38.86/16.65/35.99 & 17.61/2.79/13.38 \\
         \bf Ours &8-8-8 & \textbf{45.96/23.15/43.45} & \textbf{42.29/19.63/34.56}  & 6-6-6 & \textbf{41.00/18.41/38.51} & \textbf{34.61/12.86/27.38} \\ 
        \bottomrule

    \end{tabular}  
    \end{adjustbox}
    \caption{PTQ results of BART model on summarization tasks in terms of ROUGE 1/2/L. }
    \label{tab_ptq_summ}
    % \vspace{-8pt}
\end{table*}

% \input{tables/ptq_squad_results}
% \input{tables/ptq_squad_results}
% \subsubsection{Results on Summarization Tasks}
% \input{tables/ptq_summ_results}

\section{Conclusions and Discussions of Limitations}
In this paper, we analyze the outlier phenomenon from the inducement and clipping impact on transformer language models. Based on these, we establish an outlier suppression framework to suppress the outliers. There also remain some open problems worthy of more in-depth investigations. For example, it is valuable to systematically explore whether the conclusion in this paper benefits other fields such as computer vision. Besides, as we supplement in the Appendix that the outliers occur not only in the fine-tuned (BERT) models but also in the pre-trained ones, it's also meaningful to dive into the pre-training process for a better understanding.

% In this paper, we mainly analyze the challenge of language transformer quantization. It is valuable to systematically explore whether the conclusion in this paper benefits other fields such as computer vision. And as we mention in the Appendix that the outlier emergence involves not only the fine-tuned (BERT) models but also the pre-trained ones, diving into the pre-training process is also a profound future topic for a better understanding of the outliers.
% \paragraph{Paragraphs}

\section*{Acknowledgment}
We sincerely thank the anonymous reviewers for their serious reviews and valuable suggestions to make this better. This work was supported in part by National Natural Science Foundation of China under Grant 62022009 and Grant 61872021, and Beijing Nova Program of Science, and Technology under Grant Z191100001119050 and the Fundamental Research Funds for the Central Universities.
{
\bibliography{neurips_2022}
\bibliographystyle{unsrt}
}

\section*{Checklist}

% %%% BEGIN INSTRUCTIONS %%%
% The checklist follows the references.  Please
% read the checklist guidelines carefully for information on how to answer these
% questions.  For each question, change the default \answerTODO{} to \answerYes{},
% \answerNo{}, or \answerNA{}.  You are strongly encouraged to include a {\bf
% justification to your answer}, either by referencing the appropriate section of
% your paper or providing a brief inline description.  For example:
% \begin{itemize}
%   \item Did you include the license to the code and datasets? \answerYes{See Section~\ref{gen_inst}.}
%   \item Did you include the license to the code and datasets? \answerNo{The code and the data are proprietary.}
%   \item Did you include the license to the code and datasets? \answerNA{}
% \end{itemize}
% Please do not modify the questions and only use the provided macros for your
% answers.  Note that the Checklist section does not count towards the page
% limit.  In your paper, please delete this instructions block and only keep the
% Checklist section heading above along with the questions/answers below.
% %%% END INSTRUCTIONS %%%

\begin{enumerate}

\item For all authors...
\begin{enumerate}
  \item Do the main claims made in the abstract and introduction accurately reflect the paper's contributions and scope?
    \answerYes{}
  \item Did you describe the limitations of your work?
    \answerYes{In Discussions we leave some topics as future work.} 
  \item Did you discuss any potential negative societal impacts of your work?
    \answerNA{}
  \item Have you read the ethics review guidelines and ensured that your paper conforms to them?
    \answerYes{}
\end{enumerate}

\item If you are including theoretical results...
\begin{enumerate}
  \item Did you state the full set of assumptions of all theoretical results?
    \answerYes{}
        \item Did you include complete proofs of all theoretical results?
    \answerYes{Detailed proofs can be found in the supplementary materials.}
\end{enumerate}

\item If you ran experiments...
\begin{enumerate}
  \item Did you include the code, data, and instructions needed to reproduce the main experimental results (either in the supplemental material or as a URL)?
    \answerYes{We provide code of experiment as part of our supplementary materials.}
  \item Did you specify all the training details (e.g., data splits, hyperparameters, how they were chosen)?
    \answerYes{We defer detailed training settings in the supplementary materials.}
        \item Did you report error bars (e.g., with respect to the random seed after running experiments multiple times)?
    \answerNo{Since we comprehensively evaluate the robust generalization for various models on different datasets, it would be computationally expensive to have the error bar.}
        \item Did you include the total amount of compute and the type of resources used (e.g., type of GPUs, internal cluster, or cloud provider)?
    \answerYes{}
\end{enumerate}

\item If you are using existing assets (e.g., code, data, models) or curating/releasing new assets...
\begin{enumerate}
  \item If your work uses existing assets, did you cite the creators?
    \answerYes{}
  \item Did you mention the license of the assets?
    \answerYes{}
  \item Did you include any new assets either in the supplemental material or as a URL?
    \answerYes{}
  \item Did you discuss whether and how consent was obtained from people whose data you're using/curating?
    \answerNA{}
  \item Did you discuss whether the data you are using/curating contains personally identifiable information or offensive content?
    \answerNA{}
\end{enumerate}

\item If you used crowdsourcing or conducted research with human subjects...
\begin{enumerate}
  \item Did you include the full text of instructions given to participants and screenshots, if applicable?
    \answerNA{}
  \item Did you describe any potential participant risks, with links to Institutional Review Board (IRB) approvals, if applicable?
    \answerNA{}
  \item Did you include the estimated hourly wage paid to participants and the total amount spent on participant compensation?
    \answerNA{}
\end{enumerate}

\end{enumerate}

%\bibliographystyle{unsrtnat}
  %%% Uncomment this line and comment out the ``thebibliography'' section below to use the external .bib file (using bibtex) .
  
\clearpage  
\appendix

\section*{Appendix}
Due to the space limitation of the main paper, we will provide supplementary analysis and experimental details in the appendix, including proof of equivalent transformation in Gamma Migration, illustration of quantization challenge, more analysis of outliers, supplementary experiments to better support our observations and methods, related works and implementation details.

\section{Supplementary illustration of Gamma Migration}
\label{appendix_proof}
In this section, we first put proof of the equivalent transformation \autoref{equation_linear_transformation}. Then the detailed migration procedures of LayerNorm after the Feed Forward network (FFN) and Cross-Attention module are given. Especially, mark the LayerNorm after FFN as FFN-LN and the one after Multi-Head Attention as MHA-LN.

\subsection{Proof of equivalent transformation}
To prove \autoref{equation_linear_transformation}, we look at each element in the output of matrix multiplication. In detail, we mark the output as $\vh$.
\begin{equation}
\begin{aligned}
\vh_i&=\sum_j\mW_{i,j} \cdot (\bm \gamma_j\cdot \vx_j)\\
&=\sum_j(\bm \gamma_j\cdot \mW_{i,j}) \cdot \vx_j.
\end{aligned}
\end{equation}
Thus, for all the elements in $\vh$, we have:
\begin{equation}
\mW(\vx\odot \begin{bmatrix}
\bm \gamma_1\\
\bm \gamma_2\\
...\\
\bm \gamma_n \end{bmatrix})=(\mW \odot \begin{bmatrix}
\bm \gamma_1 & \bm \gamma_2 & ... & \bm \gamma_n \\
\bm \gamma_1 & \bm \gamma_2 & ... & \bm \gamma_n \\
...\\
\bm \gamma_1 & \bm \gamma_2 & ... & \bm \gamma_n\end{bmatrix})\vx, 
\end{equation}
% By taking $\mV_{i,j}^{(\ell)}=\vu_j^{(\ell)}$, we have
% \begin{equation}
% \begin{aligned}
% \vz_i^{(\ell+1)}
% &=\sum_j(1+\mV_{i,j}^{(\ell)})\cdot \mW_{i,j}^{(\ell)}\cdot \va_j^{(\ell)}.
% \end{aligned}
% \end{equation}
The parameter $\bm \gamma$ is shared across samples and tokens, then the above equation always holds, and the weight in the next layer can absorb the $\bm \gamma$ naturally.

\subsection{Gamma Migration on other structures}
\begin{figure}[ht]
    \centering
    \includegraphics[width=\textwidth]{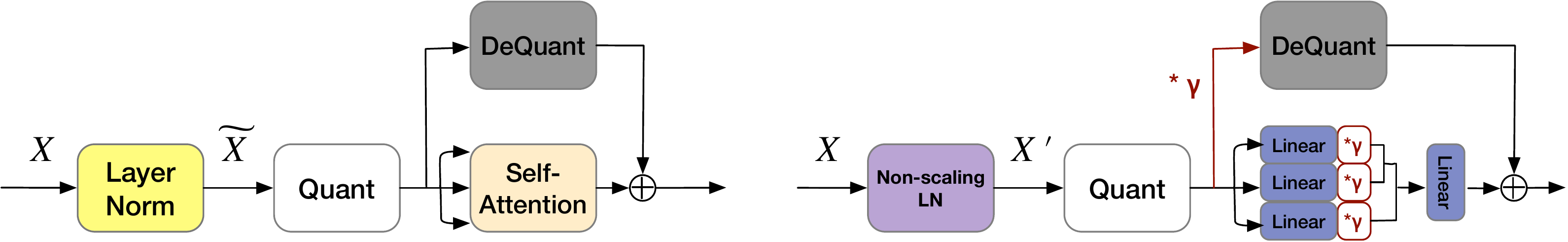}
    \caption{Comparison of the quantization flow before (left) and after (right) Gamma Migration in FFN-LN. The original LayerNorm = the Non-scaling LayerNorm * $\bm \gamma$.} 
    \label{fig_method_lnsplit_ffn}
\end{figure}
\begin{figure}[ht]
    \centering
    \includegraphics[width=\textwidth]{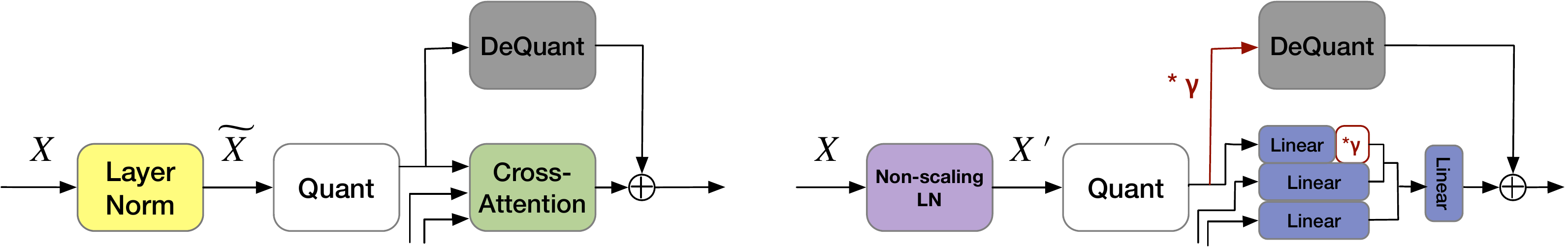}
    \caption{Comparison of the quantization flow before (left) and after (right) Gamma Migration in MHA-LN of the Cross-Attention module.} 
    \label{fig_method_lnsplit_encoder_decoder}
\end{figure}

\section{Quantization nodes}

\subsection{Position of quantization nodes}
\label{appendix_quantization_nodes}
% For quantization node which indicates the corresponding weight or activation needs to be quantized
For the position to insert quantization nodes, we find that different papers often have different choices, particularly at activation. This would bring difficulties for fair comparisons across methods and practical development on hardware. 

By surveying multiple industry~\cite{FasterTransformer, NNCFTransformer} and academic solutions, we take the one in FasterTransformer~\cite{FasterTransformer}: Token (position, token type) embeddings are quantized to reduce the memory storage. Weights and activation engaged in matrix multiplication are also quantized. To be noted, we only give one quantizer to the same activation because it is friendly to hardware. Thus we will quantize the shortcut branch and take the same quantization parameter for the input of Query, Key, and Value modules, where some papers ~\cite{bai2020binarybert} do not and might suffer some problems on hardware. 

\begin{figure}[t]
    \centering
    \includegraphics[width=\textwidth]{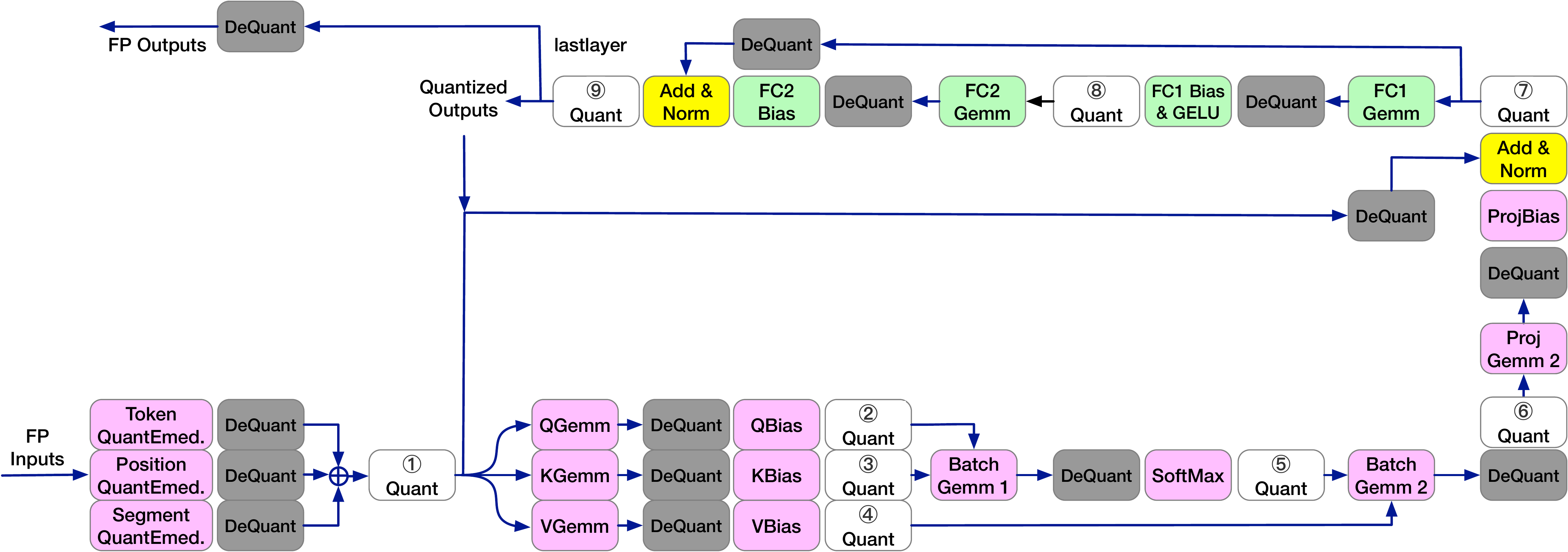}
    \caption{Position of activation quantization nodes. In real inference, the adjacent "DeQuant" and "Quant" operations will be merged into one "ReQuant" operation for faster computation.} 
    \label{fig_quantization_node}
\end{figure}

A clear illustration about the position of activation quantization is depicted in \autoref{fig_quantization_node}. Here for ease of understanding, we mark each "Quant" node with a serial number and match them with the related module names in \autoref{tab_mapping_quant2module}.

\begin{table*}[ht]
\footnotesize
    \centering
    \begin{adjustbox}{max width=\textwidth}
    \renewcommand \tabcolsep{1.0em}
    \begin{tabular}{l|l|l|l|l|l|l|l|l}
        \toprule
         \ding{172} & \ding{173} & \ding{174} & \ding{175} & \ding{176} & \ding{177} & \ding{178} & \ding{179} & \ding{180} \\
         \midrule
         Input Embedding & Query & Key & Value & Attention probs & Context & MHA-LN & GELU & FFN-LN \\
        \bottomrule
    \end{tabular}
    \end{adjustbox}
    \caption{We map the label in \autoref{fig_quantization_node} to the module name, which represents the quantization node inserted at the output of the corresponding module. }
    \label{tab_mapping_quant2module}
\end{table*}

\subsection{Problematic quantization nodes}
\label{appendix_problematic_quantization_nodes}
In this subsection, we give some simple and direct studies to elaborate on the most problematic tensors (outputs of LayerNorm structures and GELU). Verifications are done on fine-tuned BERT, RoBERTa, and encoder-decoder model BART. 

On the one hand, we compare the cosine similarity between the FP value and the quantized one for each output.
Activation nodes with cosine similarity lower than 0.99 are viewed as problematic positions (results in \autoref{tab_output_quantizer_cosine}, \autoref{tab_bart_output_quantizer_cosine}). On another hand, we can observe the final accuracy recovery by disabling the quantization of each kind of activation. Both experiments indicate the obstacles when quantizing the outputs of LayerNorm and GELU.

\begin{table*}[htp!]
\footnotesize
    \centering
    \begin{adjustbox}{max width=\textwidth}
    \begin{tabular}{lllllllllllll}
        \toprule
          \multicolumn{2}{c}{\bf BERT-STS-B} & \multicolumn{2}{c}{\bf BERT-QQP} & \multicolumn{2}{c}{\bf BERT-MRPC}\\
            \cmidrule(l{2pt}r{2pt}){1-2}   
            \cmidrule(l{2pt}r{2pt}){3-4} 
            \cmidrule(l{2pt}r{2pt}){5-6} 
        output & cosine similarity (\%) & output & cosine similarity (\%) & output & cosine similarity (\%)\\

        \midrule
layer.8.GELU &87.83 &layer.9.GELU &94.19 &layer.9.GELU &92.00\\
layer.11.GELU &90.68 &layer.4.MHA-LN &94.40 &layer.7.MHA-LN &93.05\\
layer.4.MHA-LN &94.60 &layer.6.MHA-LN &94.45 &layer.8.MHA-LN &93.14\\
layer.6.MHA-LN &94.63 &layer.5.MHA-LN &94.55 &layer.6.MHA-LN &93.22\\
layer.5.MHA-LN &94.66 &layer.7.MHA-LN &94.60 &layer.4.MHA-LN &93.28\\
layer.7.MHA-LN &94.85 &layer.3.MHA-LN &95.01 &layer.5.MHA-LN &93.44\\
layer.3.MHA-LN &95.19 &layer.8.MHA-LN &95.05 &layer.2.GELU &93.94\\
layer.10.MHA-LN &95.45 &layer.2.GELU &95.08 &layer.3.MHA-LN &94.15\\
layer.8.MHA-LN &95.45 &layer.9.MHA-LN &95.80 &layer.10.MHA-LN &94.36\\
layer.2.GELU &95.48 &layer.10.MHA-LN &96.13 &layer.9.MHA-LN &94.58\\
layer.9.MHA-LN &95.60 &layer.1.MHA-LN &96.84 &layer.8.GELU &94.68\\
layer.5.GELU &96.86 &layer.0.MHA-LN &96.87 &layer.10.GELU &95.81\\
layer.0.MHA-LN &96.96 &layer.10.GELU &97.02 &layer.0.MHA-LN &96.99\\
layer.1.MHA-LN &97.15 &layer.2.MHA-LN &97.50 &layer.1.MHA-LN &97.12\\
layer.9.GELU &97.42 &layer.4.GELU &97.57 &layer.2.MHA-LN &97.66\\
layer.4.GELU &97.60 &layer.5.GELU &97.71 &layer.11.GELU &97.70\\
layer.2.MHA-LN &97.67 &layer.3.GELU &98.30 &layer.5.GELU &97.91\\
layer.6.GELU &98.07 &layer.11.GELU &98.43 &layer.4.GELU &98.04\\
layer.3.GELU &98.22 &layer.1.GELU &98.46 &layer.11.MHA-LN &98.16\\
layer.1.GELU &98.34 &layer.0.GELU &98.60 &layer.1.GELU &98.18\\
layer.7.GELU &98.43 &layer.8.GELU &98.63 &layer.0.GELU &98.31\\
layer.10.GELU &98.44 &layer.7.GELU &98.69 &layer.7.GELU &98.42\\
layer.0.GELU &98.52 &layer.11.MHA-LN &98.76 &layer.3.GELU &98.67\\
layer.11.MHA-LN &98.60 &layer.6.GELU &98.77 &layer.6.GELU &98.74\\
layer.9.FFN-LN &98.79 &layer.10.Context &98.96 &layer.10.FFN-LN &98.94\\

        \toprule
  \multicolumn{2}{c}{\bf RoBERTa-MNLI} & \multicolumn{2}{c}{\bf RoBERTa-QNLI} & \multicolumn{2}{c}{\bf RoBERTa-QQP}\\
    \cmidrule(l{2pt}r{2pt}){1-2}   
    \cmidrule(l{2pt}r{2pt}){3-4} 
    \cmidrule(l{2pt}r{2pt}){5-6} 
    output & cosine similarity(\%) & output & cosine similarity(\%) & output & cosine similarity(\%)\\
    \midrule
 layer.7.GELU &93.91 &layer.10.GELU &90.08 &layer.2.GELU &93.56\\
layer.9.GELU &94.25 &layer.7.GELU &91.60 &layer.3.GELU &94.27\\
layer.2.GELU &94.64 &layer.5.GELU &95.58 &layer.4.GELU &95.96\\
layer.10.GELU &94.79 &layer.4.GELU &95.59 &layer.1.GELU &96.69\\
layer.8.GELU &94.83 &layer.2.GELU &95.89 &layer.5.GELU &96.71\\
layer.5.GELU &96.16 &layer.8.GELU &96.02 &layer.0.GELU &97.04\\
layer.4.GELU &96.28 &layer.3.GELU &96.33 &layer.0.MHA-LN &97.09\\
layer.1.GELU &96.38 &layer.1.GELU &96.52 &layer.7.GELU &97.41\\
layer.3.GELU &96.69 &layer.9.GELU &96.85 &layer.1.MHA-LN &97.59\\
layer.6.GELU &96.82 &layer.11.MHA-LN &97.00 &layer.8.GELU &97.81\\
layer.0.MHA-LN &97.16 &layer.0.MHA-LN &97.13 &layer.8.FFN-LN &98.10\\
layer.11.MHA-LN &97.26 &layer.6.GELU &97.36 &layer.7.FFN-LN &98.13\\
layer.0.GELU &97.30 &layer.0.GELU &97.49 &layer.0.FFN-LN &98.16\\
layer.10.FFN-LN &97.64 &layer.1.MHA-LN &97.66 &layer.1.FFN-LN &98.23\\
layer.10.MHA-LN &97.64 &layer.8.Context &97.67 &layer.6.FFN-LN &98.28\\
layer.1.MHA-LN &97.67 &layer.10.FFN-LN &97.72 &layer.6.GELU &98.29\\
layer.9.FFN-LN &97.84 &layer.10.MHA-LN &97.75 &layer.7.MHA-LN &98.32\\
layer.8.FFN-LN &97.90 &layer.9.Context &97.79 &layer.8.MHA-LN &98.33\\
layer.7.FFN-LN &98.05 &layer.9.FFN-LN &97.89 &layer.6.MHA-LN &98.35\\
layer.9.MHA-LN &98.11 &layer.8.FFN-LN &97.92 &layer.5.FFN-LN &98.36\\
layer.8.MHA-LN &98.13 &layer.7.FFN-LN &97.99 &layer.2.MHA-LN &98.42\\
layer.0.FFN-LN &98.14 &layer.0.FFN-LN &98.14 &layer.5.MHA-LN &98.43\\
layer.6.FFN-LN &98.25 &layer.8.MHA-LN &98.15 &layer.4.FFN-LN &98.46\\
layer.1.FFN-LN &98.33 &layer.9.MHA-LN &98.17 &layer.3.MHA-LN &98.49\\
layer.5.FFN-LN &98.34 &layer.6.FFN-LN &98.19 &layer.4.MHA-LN &98.50\\
layer.6.MHA-LN &98.36 &layer.5.FFN-LN &98.26 &layer.2.FFN-LN &98.50\\
layer.7.MHA-LN &98.36 &layer.6.MHA-LN &98.28 &layer.9.FFN-LN &98.52\\
layer.4.FFN-LN &98.39 &layer.7.MHA-LN &98.31 &layer.3.FFN-LN &98.57\\
layer.5.MHA-LN &98.43 &layer.1.FFN-LN &98.32 &layer.10.GELU &98.58\\
layer.4.MHA-LN &98.47 &layer.4.FFN-LN &98.34 &layer.9.MHA-LN &98.60\\
layer.3.FFN-LN &98.48 &layer.5.MHA-LN &98.37 &layer.10.FFN-LN &98.60\\
layer.2.MHA-LN &98.50 &layer.3.FFN-LN &98.45 &layer.11.MHA-LN &98.75\\
layer.2.FFN-LN &98.54 &layer.4.MHA-LN &98.45 &layer.9.GELU &98.86\\
layer.3.MHA-LN &98.55 &layer.2.FFN-LN &98.50 &layer.10.MHA-LN &98.89\\
 & &layer.2.MHA-LN &98.52 & &\\
 & &layer.3.MHA-LN &98.55 & &\\
    \bottomrule
    \end{tabular}
    \end{adjustbox}
    \caption{The sorted cosine similarity between the output and the quantized one (6-bit) on BERT and RoBERTa models. We aim at the most problematic ones with cosine similarity lower than 99\%. }
    \label{tab_output_quantizer_cosine}
\end{table*}
\clearpage
\begin{table*}[htp!]
\footnotesize
    \centering
    \begin{adjustbox}{max width=\textwidth}
    \begin{tabular}{lllllllllllll}
        \toprule
          \multicolumn{2}{c}{\bf BART-CNN/DailyMail} & \multicolumn{2}{c}{\bf BART-XSum}\\
            \cmidrule (l{2pt}r{2pt}){1-2}   
            \cmidrule (l{2pt}r{2pt}){3-4} 
        % & similarity  (\%)  &  & similarity  (\%)  &  & similarity  (\%) &  & similarity  (\%) &  & similarity  (\%)\\
        output & cosine similarity (\%) & output & cosine similarity (\%)\\
        \midrule
        layers.4.GELU (Decoder) &67.96 &layers.3.GELU (Decoder) &74.37\\
        layers.3.GELU (Decoder) &69.50 &layers.4.GELU (Decoder) &75.05\\
        layers.4.MHA-LN (Encoder-Decoder) &76.03 &layers.2.GELU (Decoder) &82.36\\
        layers.2.GELU (Decoder) &76.05 &layers.4.MHA-LN (Encoder-Decoder) &82.84\\
        layers.2.MHA-LN (Encoder-Decoder) &77.88 &layers.1.MHA-LN (Encoder) &83.04\\
        layers.0.GELU (Decoder) &80.83 &layers.2.MHA-LN (Encoder-Decoder) &84.31\\
        layers.5.MHA-LN (Encoder) &84.20 &layers.4.MHA-LN (Encoder) &84.53\\
        layers.1.MHA-LN (Encoder) &84.33 &layers.5.MHA-LN (Encoder) &84.69\\
        layers.1.MHA-LN (Encoder-Decoder) &85.01 &layers.3.MHA-LN (Encoder) &86.47\\
        layers.4.MHA-LN (Encoder) &85.03 &layers.1.MHA-LN (Encoder-Decoder) &86.97\\
        layers.3.MHA-LN (Encoder-Decoder) &86.78 &layers.0.MHA-LN (Encoder) &87.69\\
        layers.3.MHA-LN (Encoder) &87.12 &layers.0.GELU (Decoder) &87.77\\
        layers.0.MHA-LN (Encoder) &87.30 &layers.3.MHA-LN (Encoder-Decoder) &88.11\\
        layers.1.GELU (Decoder) &87.61 &layers.2.MHA-LN (Encoder) &89.14\\
        layers.2.MHA-LN (Encoder) &89.64 &layers.0.GELU (Encoder) &92.21\\
        layers.5.GELU (Decoder) &91.78 &layers.1.GELU (Decoder) &93.60\\
        layers.0.MHA-LN (Encoder-Decoder) &93.62 &layers.0.MHA-LN (Encoder-Decoder) &93.61\\
        layers.0.GELU (Encoder) &95.09 &layers.5.FFN-LN (Decoder) &95.44\\
        layers.2.GELU (Encoder) &95.91 &layers.5.GELU (Decoder) &96.35\\
        layers.3.GELU (Encoder) &96.44 &layers.3.GELU (Encoder) &96.41\\
        layers.3.MHA-LN (Decoder) &96.90 &layers.2.GELU (Encoder) &96.57\\
        layers.5.MHA-LN (Decoder) &97.46 &layers.3.MHA-LN (Decoder) &96.87\\
        layers.2.Context (Encoder-Decoder) &97.51 &layers.2.Context (Encoder-Decoder) &96.99\\
        layers.5.MHA-LN (Encoder-Decoder) &97.71 &layers.1.GELU (Encoder) &97.20\\
        layers.4.FFN-LN (Decoder) &97.83 &layers.5.MHA-LN (Encoder-Decoder) &97.56\\
        layers.4.GELU (Encoder) &97.85 &layers.0.Context (Encoder-Decoder) &97.72\\
        layers.1.GELU (Encoder) &97.88 &layers.5.GELU (Encoder) &97.74\\
        layers.5.GELU (Encoder) &97.97 &layers.4.FFN-LN (Decoder) &98.02\\
        layers.5.FFN-LN (Decoder) &98.32 &layers.4.GELU (Encoder) &98.04\\
        layers.2.Context (Decoder) &98.40 &layers.0.Context (Decoder) &98.11\\
        layers.1.MHA-LN (Decoder) &98.51 &layers.5.MHA-LN (Decoder) &98.20\\
        layers.3.FFN-LN (Decoder) &98.52 &layers.2.FFN-LN (Decoder) &98.28\\
        layers.0.Context (Decoder) &98.53 &layers.3.Context (Encoder-Decoder) &98.31\\
        layers.4.MHA-LN (Decoder) &98.54 &layers.1.Context (Decoder) &98.32\\
        layers.2.MHA-LN (Decoder) &98.63 &layers.3.FFN-LN (Decoder) &98.36\\
        layers.2.FFN-LN (Decoder) &98.66 &layers.1.MHA-LN (Decoder) &98.38\\
        layers.1.FFN-LN (Decoder) &98.71 &layers.5.Context (Encoder-Decoder) &98.46\\
        layers.0.Context (Encoder-Decoder) &98.71 &layers.2.Context (Decoder) &98.56\\
        layers.0.FFN-LN (Decoder) &98.72 &layers.4.MHA-LN (Decoder) &98.58\\
        layers.5.Context (Encoder-Decoder) &98.72 &layers.2.MHA-LN (Decoder) &98.64\\
        layers.4.Context (Decoder) &98.92 &layers.0.FFN-LN (Decoder) &98.71\\
        layers.0.MHA-LN (Decoder) &98.93 &layers.1.FFN-LN (Decoder) &98.72\\
         &  &layers.0.MHA-LN (Decoder) &98.80\\
        \bottomrule
    \end{tabular}
    \end{adjustbox}
    \caption{The sorted cosine similarity between the output and the quantized one (6-bit) on BART models. We aim at the most problematic ones with cosine similarity lower than 99\%.}
    \label{tab_bart_output_quantizer_cosine}
\end{table*}

\begin{table*}[htp!]
\footnotesize
    \centering
    \begin{adjustbox}{max width=\textwidth}
    \begin{tabular}{lcc|cccccccccc}
        \toprule
            \bf Model & 32-32-32 & 6-6-6 & Input Embedding & Query & Key & Value & Attention probs & Context & MHA-LN & GELU & FFN-LN \\
        \midrule
        BERT-MRPC & 87.75 & 31.86 & 31.62 & 32.11 & 32.11 & 32.6 & 31.62 & 31.86 & \bf 83.09 & \bf 34.31 & 31.86 \\
        BERT-QQP & 90.91 & 69.0 &69.22 & 69.05 & 68.95 & 69.24 & 68.09 & 69.25 & \bf 88.93 & \bf 74.25 & 70.01 \\
        BERT-STS-B & 89.70 & 59.79 & 57.8 & 57.61 & 58.2 &56.45 & 54.02 & 55.12 & \bf 84.1 & \bf 79.6 & 53.68 \\
        % BERT-RTE & & 47.65 & 49.1 & 47.65 & 48.01 & 47.65 & 47.29 & \bf 61.37 & \bf 55.6 & 47.29 & 47.65 & 47.29\\
        \midrule
        RoBERTa-MNLI & 87.75  & 34.90 & 36.05  & 35.69 & 35.54 & 35.27 &  35.68 & 36.08 & \bf 66.93 & \bf 60.50 & \bf 53.82\\
        RoBERTa-QNLI  & 92.68 & 62.13 & 65.04 & 65.77 & 64.23 & 64.73 & 64.54 & 64.42 & \bf 84.55 & \bf 69.71 & \bf 76.66 \\
        RoBERTa-QQP & 91.6 & 74.37 & 76.24 & 75.97 & 76.01 & 76.41 & 75.50& 75.92 & \bf 87.80& \bf 84.28 & \bf80.89  \\
        \bottomrule
    \end{tabular}
    \end{adjustbox}
\caption{Influence study of quantization nodes. The comparisons of the second and third columns show the performance drop with 6-bit MinMax calibration and quantization. The subsequent columns show the recovered performance after disabling the quantization of a certain kind of output defined in \autoref{tab_mapping_quant2module}, which implies the effect of quantizing this node. For example, "Query" means disabling the quantization of output at Query modules across 12 layers. Obvious improvements are marked in bold.}
    \label{tab_output_quantizer_accuracy}
\end{table*}

\section{Analysis of outliers}
\subsection{Outlier phenomenon}
\label{appendix_outliers phenomenon}
\begin{figure}[t]
%sst2 attn6 layernorm
    \centering
    \includegraphics[width=0.8\textwidth]{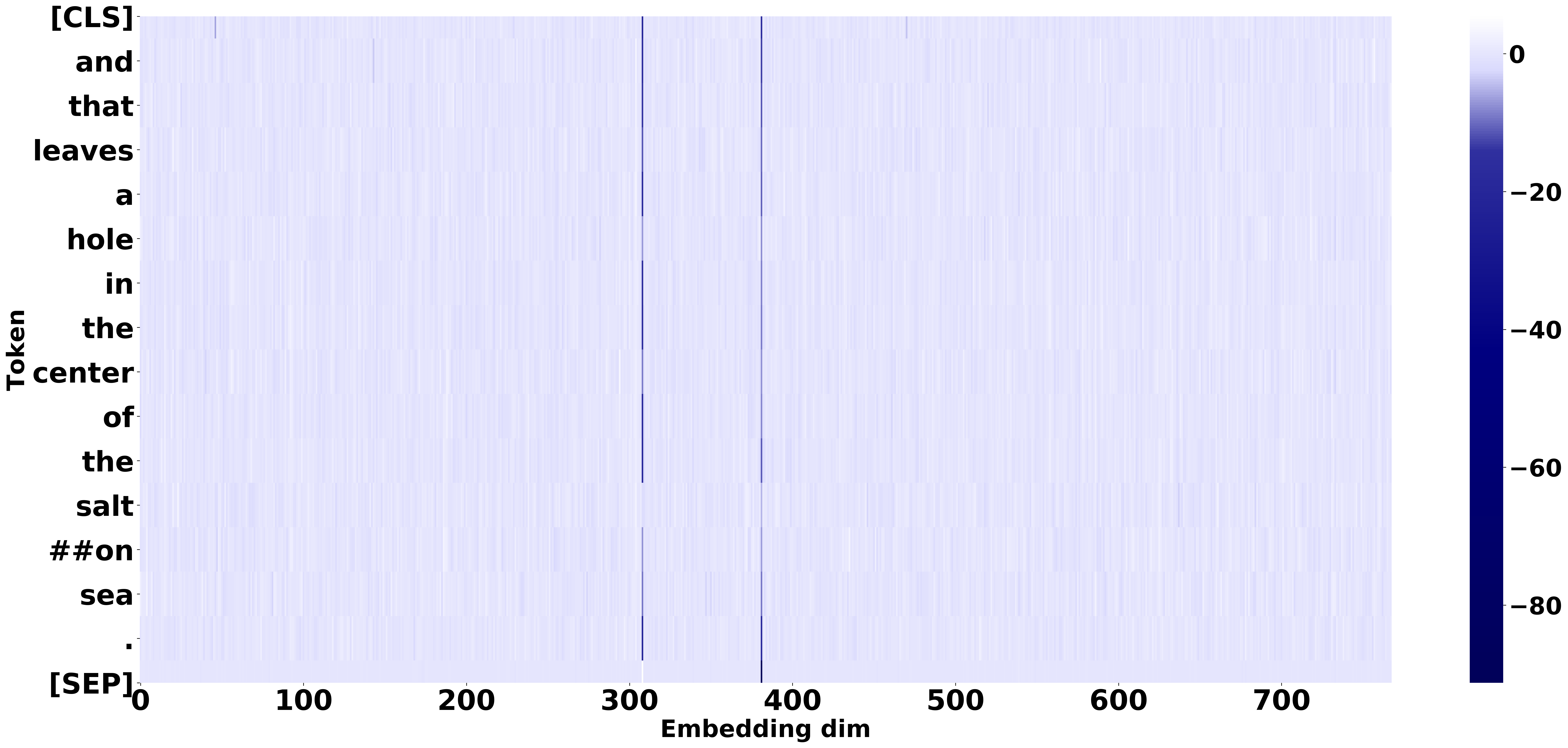}
    \caption{The dark strips on embedding dim 308 and 381 represent the outliers across almost all tokens at LayerNorm's output in BERT-SST-2. }
    \label{fig_embedding_outliers}
\end{figure}

\begin{figure}[b]
%sst2 attn4 layernorm
    \centering
    \includegraphics[width=0.8\textwidth]{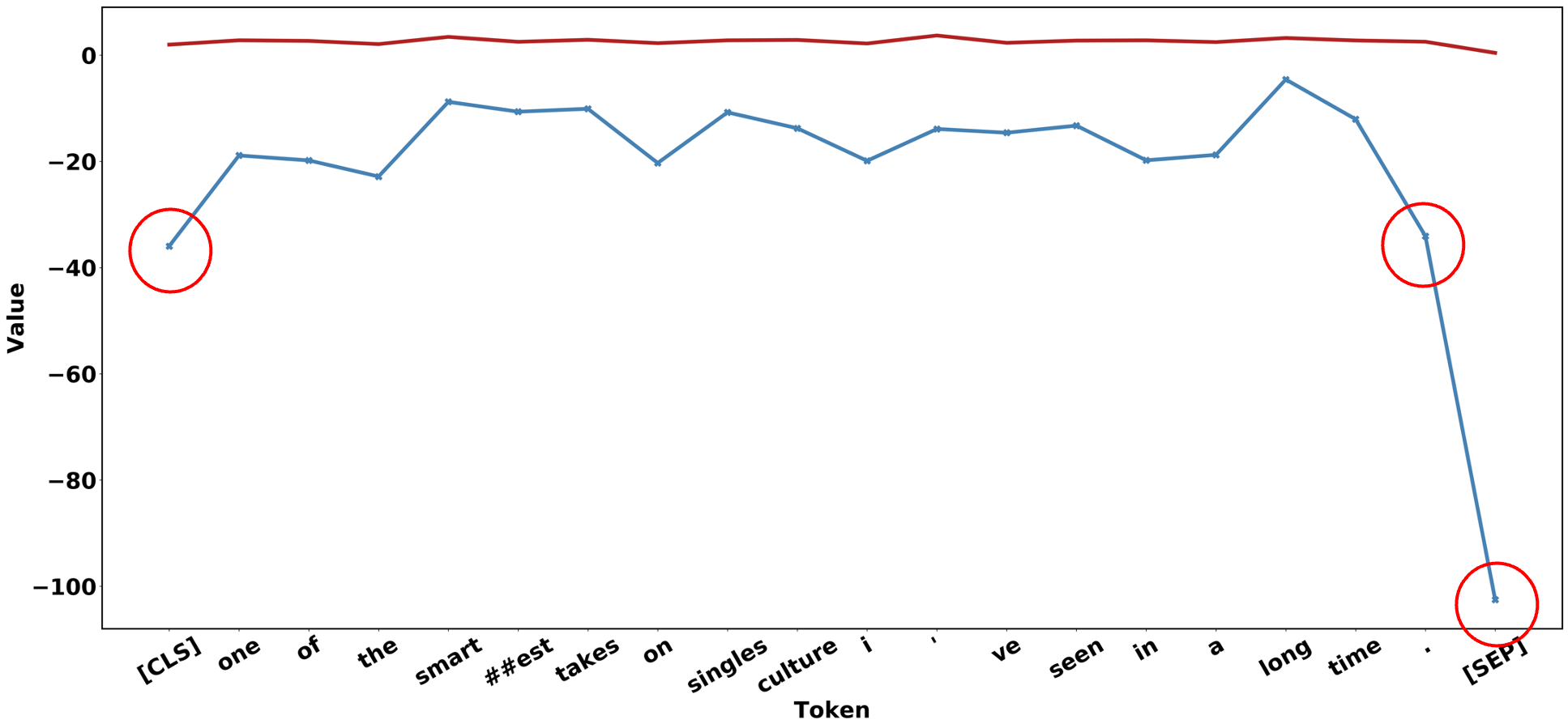}
    \caption{We draw the token range and token [SEP], [CLS], '.' attend to sharper outliers here as marked in red circles.} 
    \label{fig_token_outliers}
\end{figure}

By going deeper into the above problematic activations, we find that large outliers in them cause the large quantization error, and these outliers present some structured features from the embedding and token perspectives. Activations of almost all tokens attend to outliers in specific embedding dimensions like 308 and 381 embedding dimensions in \autoref{fig_embedding_outliers}. Upon these dimensions, some tokens like the [SEP] token in \autoref{fig_token_outliers} attend to even more aggressive outliers compared to other tokens in (\autoref{fig_token_outliers}). In fact, we find this often happens on token [SEP], [CLS], punctuations like commas and periods, and other high-frequency tokens like "the", "and", "of".

\subsection{Detailed discussion about the inducement}
\label{appendix_analysis_outlier}
Here, we discuss the inducement of the outlier phenomenon from embedding and token perspectives. 

For the embedding phenomenon, the \autoref{subsec_observation1} has explained the scaling parameter amplifies the outliers at certain embeddings. In fact, we find that this not only emerges in fine-tuned models but is also obvious in the pre-trained ones. By injecting constraints such as weight decay or kurtosis regularization~\cite{kure} to LayerNorm's parameter when fine-tuning the FP model, it is still hard to suppress the aggressive values in the scaling parameter without affecting FP performance. Hence, we conjecture that this phenomenon is beneficial to the FP performance though it indeed brings challenges to quantization. 

Moreover, the huge deviation in the token range we think is caused by the token frequency in the pre-training phase.  Because we find the tokens which hold more aggressive signals occur frequently during pre-training like [SEP], [CLS] occur in each example, and '.' is often used in an expression. We also notice that these tokens’ word (token) embeddings have larger values than others. According to these, a possible explanation might be like: the frequency information biases the word embedding space and brings different features. The sharper outliers spread to subsequent layers and seem to be less important as indicated in \autoref{subsec_observation2}. Therefore, we conjecture that a good word embedding without being biased by frequency information can behave better in quantization. But we can find those less important outliers in an efficient way and clip them as well. This suits better for post-training quantization without large-scale re-training.

For the inducement of outliers, note that \cite{bert_busters} also mentioned the connection between the scaling parameter and outliers in the last LayerNorm each BERT layer. But we emphasize the amplification effect of the scaling parameter, especially for the LayerNorm after Multi-Head Attention. This naturally generates the finding of quantization-friendly distribution contributed by removing the scaling parameter. About the unbalanced token frequency, a concurrent work~\cite{token_frequency} explores carefully from the FP performance perspective.

\section{Supplementary experiments}
\subsection{Supplementary evidence of outliers in LayerNorm}
\label{appendix_outlier_inducement}
\begin{figure}[b]
    \begin{subfigure}[t]{0.3\textwidth}
        \centering
        \includegraphics[height=\textwidth]{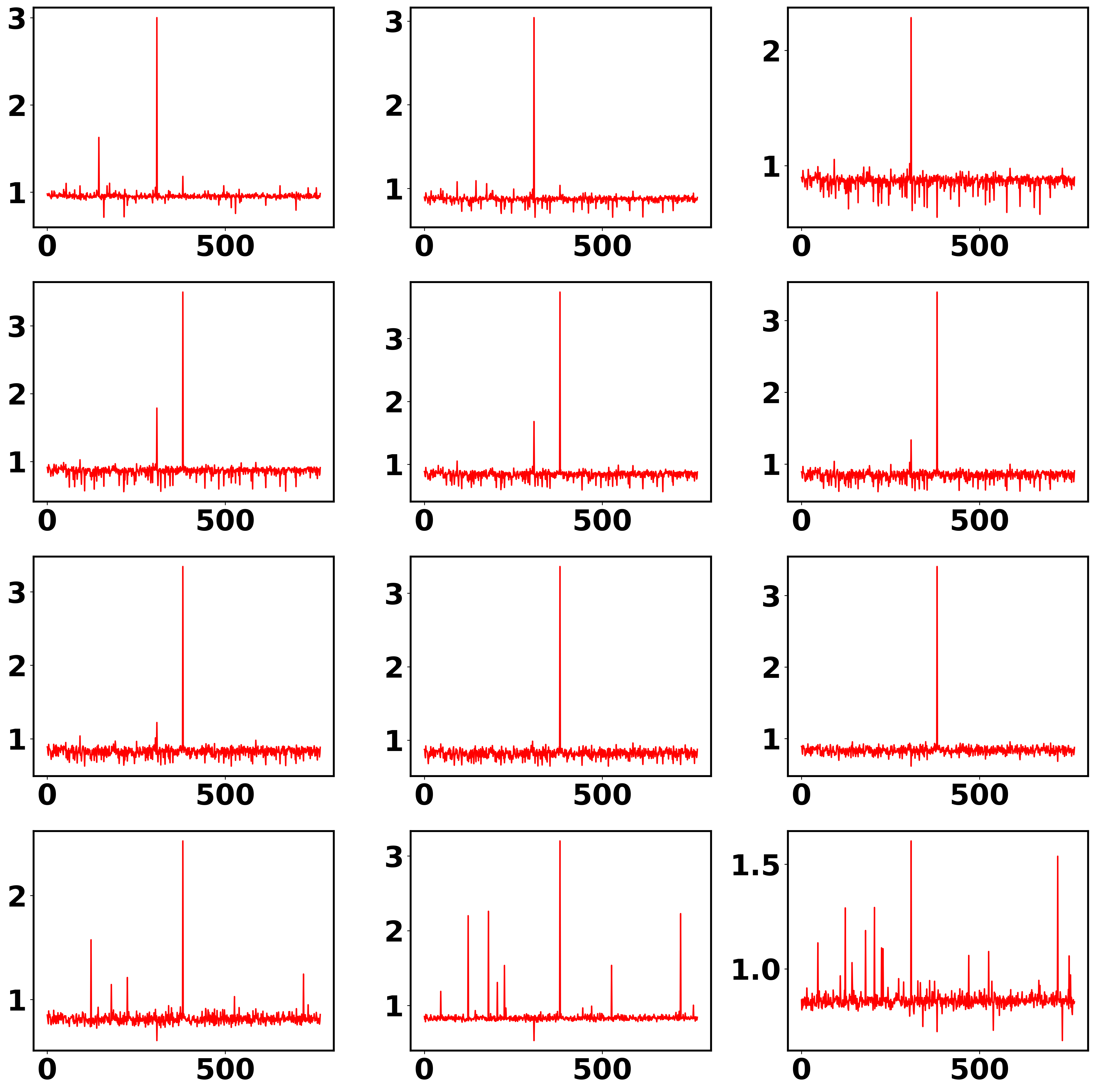}
        \caption{$\bm \gamma$}
    \end{subfigure}
    \begin{subfigure}[t]{0.3\textwidth}
        \centering
        \includegraphics[height=\textwidth]{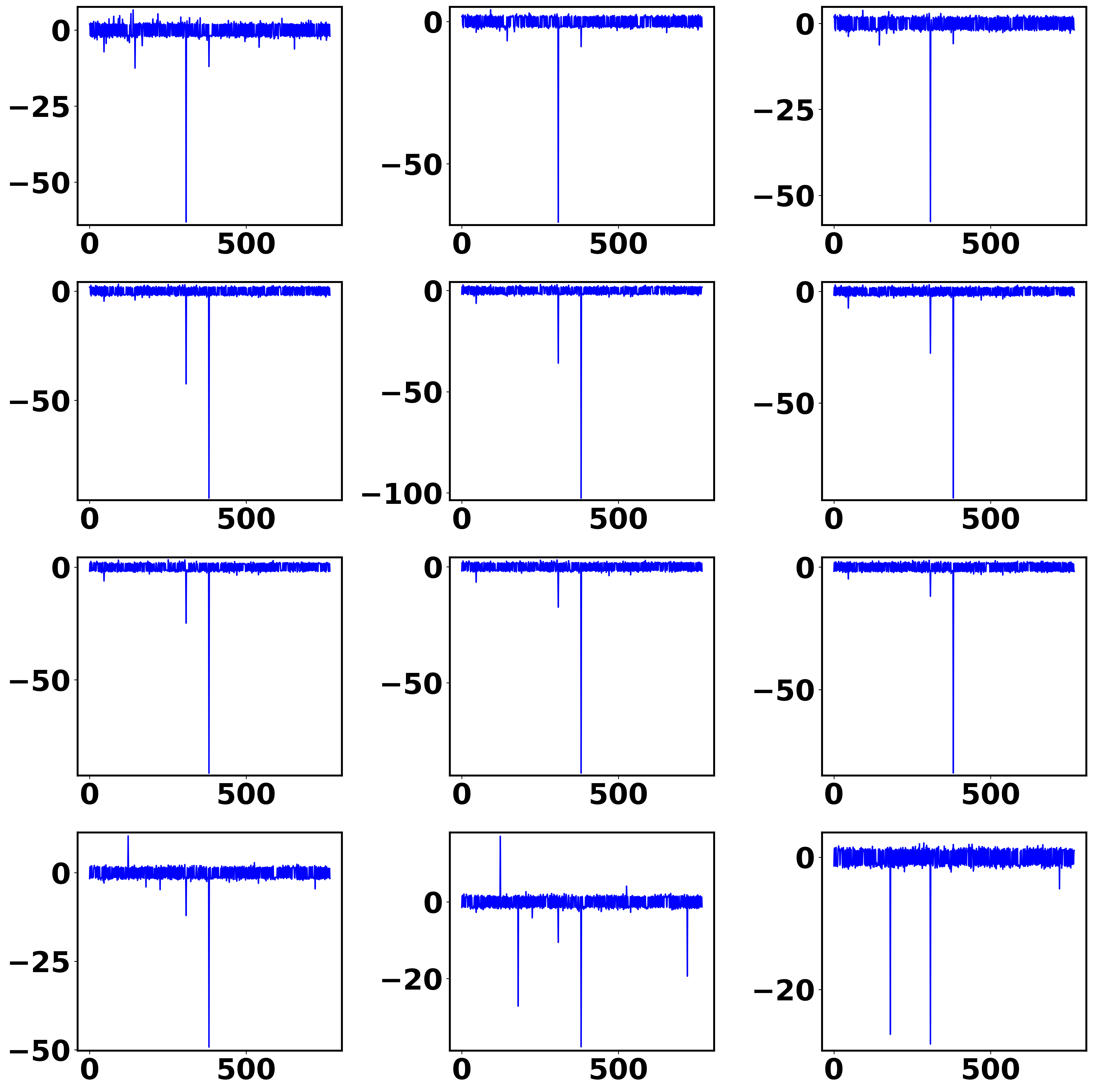}
        \caption{$\widetilde{\mX}$}
    \end{subfigure}
    \begin{subfigure}[t]{0.3\textwidth}
        \centering
        \includegraphics[height=\textwidth]{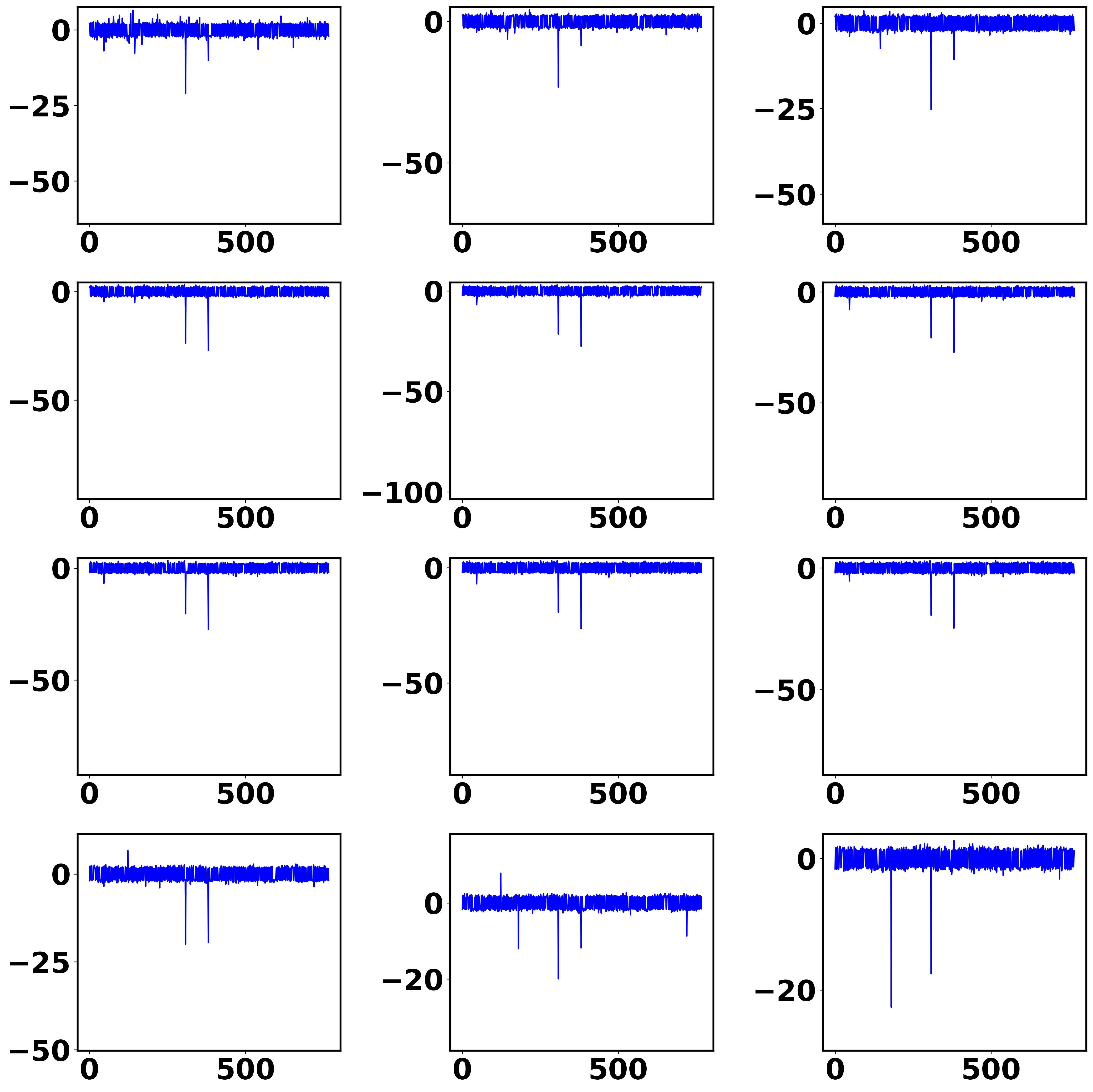}
        \caption{$\mX'$}
    \end{subfigure}
    \caption{$\bm \gamma$, $\widetilde{\mX}$ and $\mX'$ across 12 MHA-LN in BERT-SST-2, where $\widetilde{\mX}=\bm \gamma \odot \mX'$. For the latter two, we draw the highest-magnitude value at each embedding dim. It can be seen that $\mX'$ holds milder distribution.}
    \label{fig_appendix_motivation1_eg1}
\end{figure}

\begin{figure}[t]
    \begin{subfigure}[t]{0.3\textwidth}
        \centering
        \includegraphics[height=0.5\textwidth]{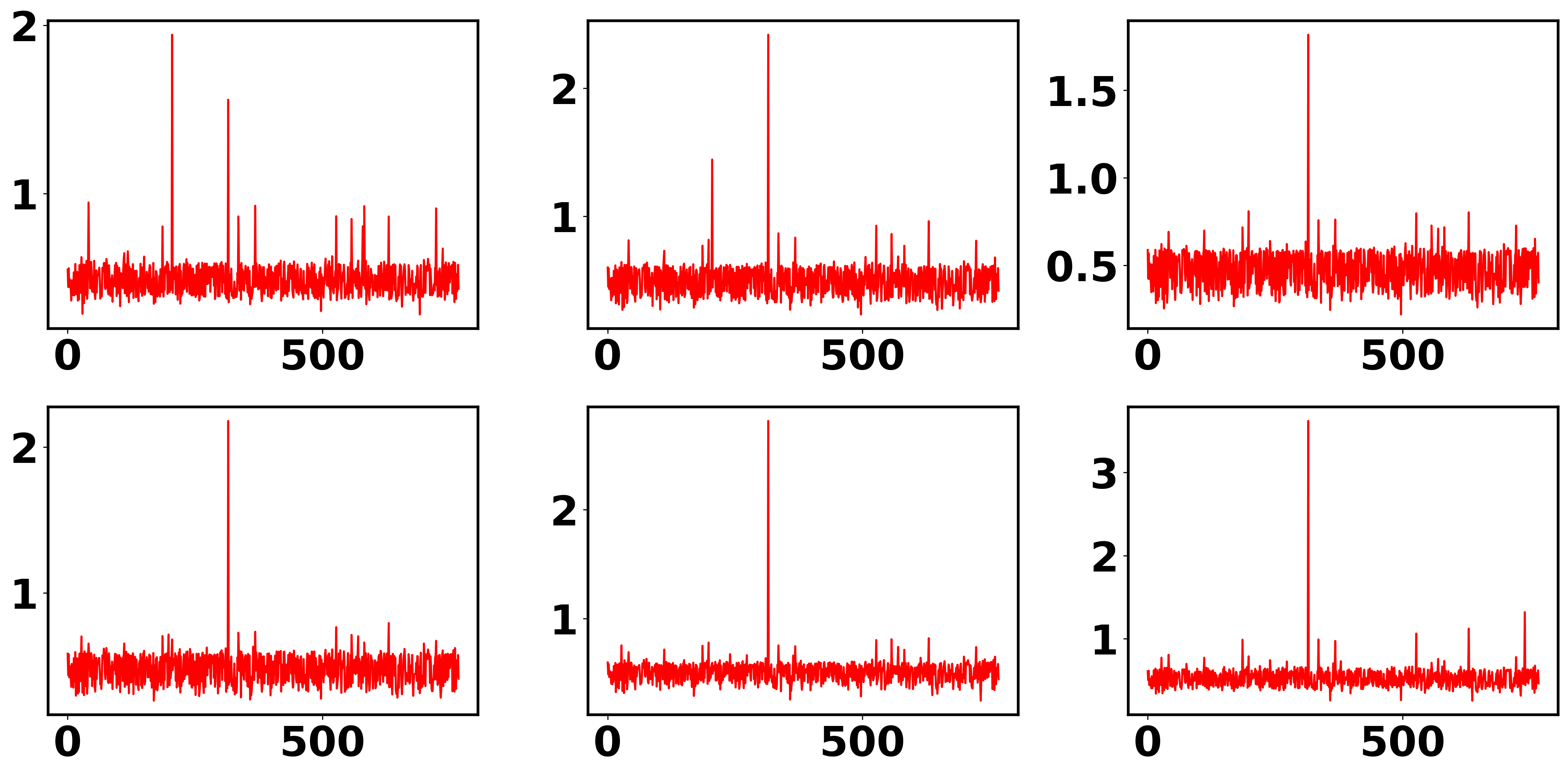}
        \caption{$\bm \gamma$}
    \end{subfigure}
    \begin{subfigure}[t]{0.3\textwidth}
        \centering
        \includegraphics[height=0.5\textwidth]{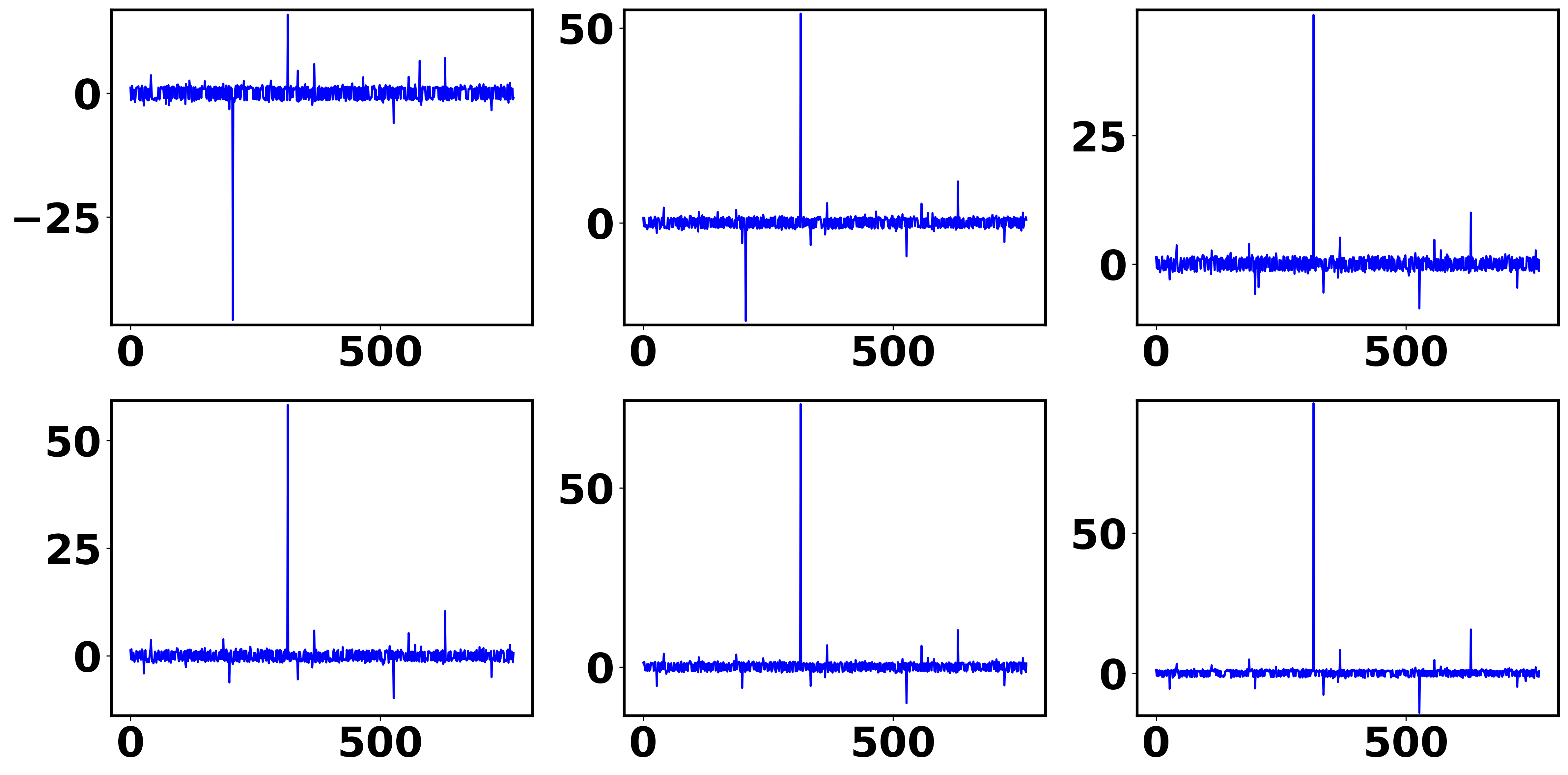}
        \caption{$\widetilde{\mX}$}
    \end{subfigure}
    \begin{subfigure}[t]{0.3\textwidth}
        \centering
        \includegraphics[height=0.5\textwidth]{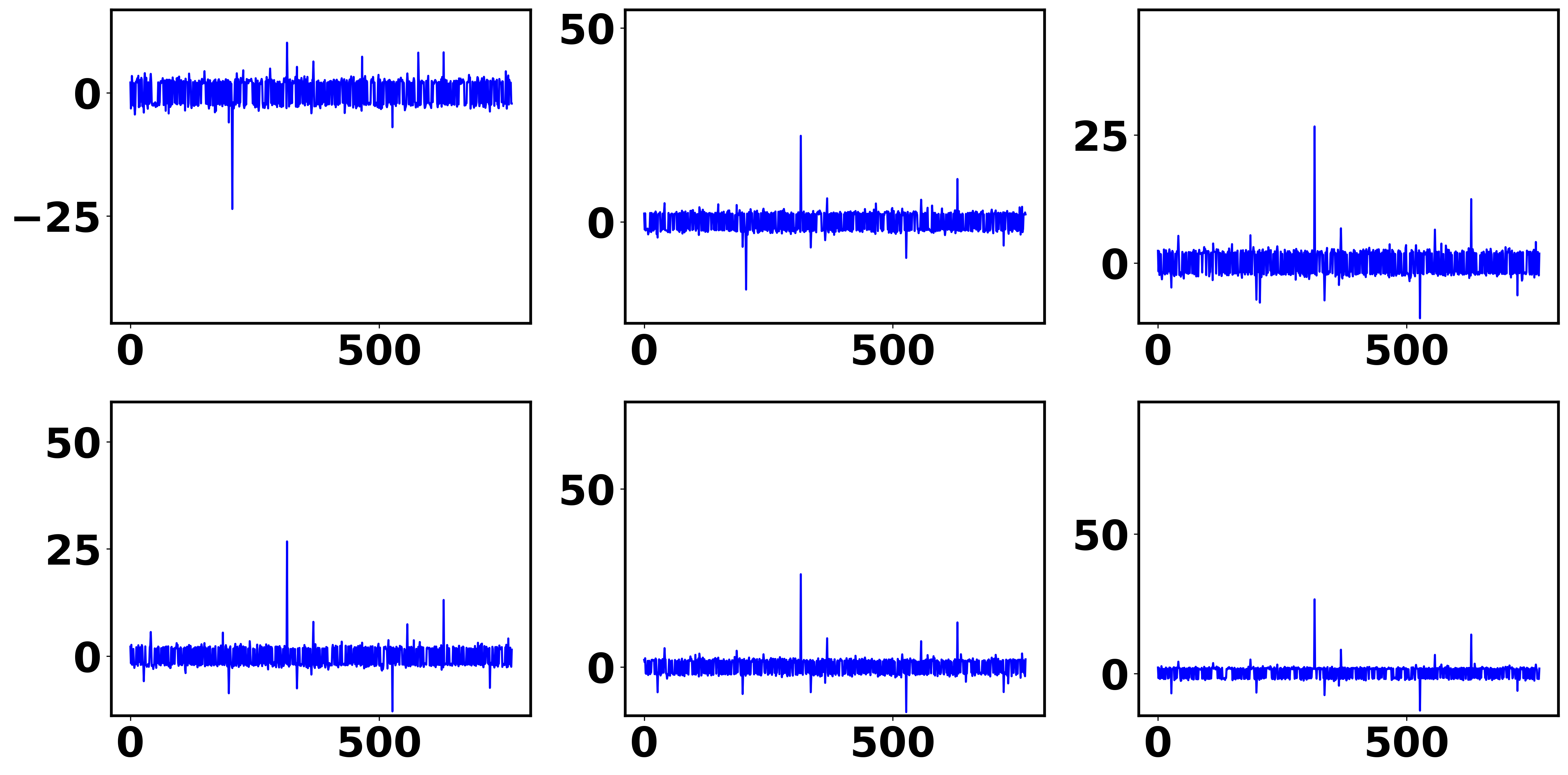}
        \caption{$\mX'$}
    \end{subfigure}
    \caption{$\bm \gamma$, $\widetilde{\mX}$ and $\mX'$ across 6 LayerNorm in BART-QQP, where $\widetilde{\mX}=\bm \gamma \odot \mX'$. For the latter two, we draw the highest-magnitude value at each embedding dim.}
     \label{fig_appendix_motivation1_eg2}
\end{figure}

We show more evidence of the same outlier phenomenon in LayerNorm and illustrate that the output of Non-scaling LayerNorm is more quantization-friendly than the normal one. Firstly, \autoref{fig_appendix_motivation1_eg1} and \autoref{fig_appendix_motivation1_eg2} are presented to build a formal understanding, where the $X'$ has weaker outliers. Furthermore, more quantitative results about cosine similarity are put in \autoref{tab_appendix_cosine_improvement} to indicate the improvement on the most problematic tensors \autoref{appendix_problematic_quantization_nodes} brought by extracting the scaling parameter $\bm \gamma$. Here, we discuss the inducement of the outlier phenomenon from embedding and token perspectives. 

% The higher cosine similarity shows that activation with extracting $\bm \gamma$ produce more availability not only to MHA-LN but also to FFN-LN and the cross-attention module in encoder-decoder model. 
% Finally, we equip our Gamma Migration proposed in \autoref{subsection_gamma_migration} to demonstrate the performance improvement directly by removing the amplifier in LayerNorm. We put the results in \autoref{}.

\begin{table*}[htp!]
\footnotesize
    \centering
    \begin{adjustbox}{max width=\textwidth}
    \begin{tabular}{lcccccccccccc}
        \toprule
          \bf Model & \textbf{0} & \textbf{1} & \textbf{2} & \textbf{3} & \textbf{4} & \textbf{5} & \textbf{6} & \textbf{7} & \textbf{8} & \textbf{9} & \textbf{10} & \textbf{11} \\
        \midrule
            BERT-MRPC \\
            \hspace{0.5em} MHA-LN & +2.24 & +2.17 & +1.48 & +4.84 & +5.70 & +5.55 & +5.76 & +5.83 & +5.56 & +4.32 & +4.79 & +0.85 \\
        \midrule
            BERT-QQP \\
            \hspace{0.5em} MHA-LN & +2.35 & +2.35 & +1.61 & +4.00 & +4.58 & +4.43 & +4.51 & +4.24 & +3.64 & + 3.07 & +3.19 & +0.24 \\
        \midrule
            BERT-STS-B \\
            \hspace{0.5em} MHA-LN & +2.19 & +2.03 & +1.43 & +3.82 & +4.39 & +4.34 & +4.35 & +4.03 & +3.25 & +3.33 & +3.44 & +0.51\\
        \midrule
            RoBERTa-MNLI \\
            \hspace{0.5em} MHA-LN & +1.49 & +0.81 & +0.25 & +0.18 & +0.16 &+ 0.16 & +0.22 & +0.19 & +0.25 &+ 0.31 & +0.59 & +1.17\\
            \hspace{0.5em} FFN-LN & +0.31 & + 0.43 & +0.16 & +0.24 & +0.25 & +0.27 & +0.28 & +0.31 & +0.34 & +0.43 & +0.49 & +0.04\\
        %     RoBERTa-QQP (4-4-4)\\
        %   \hspace{0.5em} MHA-LN & +15.37 & +4.12 & +0.30 & +1.50 & +0.69 & +0.53 & +1.39 & +3.28 & +5.72 & +4.63 & +4.64& +7.75 \\
        %   \hspace{0.5em} FFN-LN & +1.42 & +1.22 & +1.40 & +1.21 & +0.38 & +0.57 & +0.54 & 
        %   +3.85 & +5.65 & +2.81 & + 5.03 & +4.19\\
        \midrule
          RoBERTa-QNLI \\
           \hspace{0.5em} MHA-LN & +1.62 & +0.88 & +0.25 & +0.19 & +0.17 & +0.18 & +0.22 & +0.18 & +0.23 & +0.24 & +0.52 & +1.31\\
           \hspace{0.5em} FFN-LN & +0.33 & +0.47 & +0.22 & +0.25 & +0.26 & 
           +0.30 & +0.28 & +0.32 & +0.31 & +0.36 & +0.49 & +0.53\\
        \midrule
        RoBERTa-QQP \\
            \hspace{0.5em} MHA-LN & +1.57 &  +0.93 & +0.32 & +0.25 & +0.21 & +0.22 & +0.29  &  +0.33 & +0.43 & + 0.39 & +0.30 & +0.64 \\
            \hspace{0.5em} FFN-LN & +0.32 & +0.52 & +0.16 & +0.24 & +0.27 & +0.33 & +0.33  &  +0.42 & +0.49 & +0.33 & +0.45 & +0.20 \\
        \midrule
        BART-CNN/DailyMail\\
        \hspace{0.5em} MHA-LN (Encoder) & +11.26 & +14.07 & +8.81 & +11.25 & +13.86 & +14.13 &  &  &  &  &  &  \\
        \hspace{0.5em} MHA-LN (Decoder) & +0.23 & +0.19 & +0.01 & +1.69 & +0.23 & +1.29 &  &  &  &  &  &   \\
        \hspace{0.5em} MHA-LN (Encoder-Decoder) & +5.21 & +13.82 & +20.94 & +11.94 & +22.74 & +1.04 &  &  &  &  &  & \\
        \hspace{0.5em} FFN-LN (Decoder) & +0.20 & +0.14 & +0.03 & +0.17 & +0.04 & +0.21 &  &  &  &  &  &  \\
        \midrule
        BART-XSum\\
        \hspace{0.5em} MHA-LN (Encoder) & +10.90 & +15.07 & +9.17 & +11.77 & +13.81 & +13.58 &  &  &  &  &  &  \\
        \hspace{0.5em} MHA-LN (Decoder) & +0.15 & +0.12 & +0.09 & +1.63 & +0.23 & +0.61 &  &  &  &  &  &   \\
        \hspace{0.5em} MHA-LN (Encoder-Decoder) & +5.09 & +11.75 & +14.50 & +10.57 & +15.96 & +1.32 &  &  &  &  &  & \\
        \hspace{0.5em} FFN-LN (Decoder) & +0.21 & +0.04 & +0.34 & +0.23 & +0.54 & +0.38 &  &  &  &  &  &  \\
        \bottomrule
    \end{tabular}
    \end{adjustbox}
    \caption{Cosine similarity (\%) improvement after extracting $\bm \gamma$ in LayerNorm. This metric is evaluated on 256 samples from dev set. (BART only has 6 layers and thus the right half is left empty. )}
    \label{tab_appendix_cosine_improvement}
    % \vspace{-8pt}
\end{table*}

\subsection{Supplementary evidence of clipping impact}
\label{appendix_clipping_value}
\begin{table*}[ht]
\footnotesize
    \centering
    \begin{adjustbox}{max width=\textwidth}
    \begin{tabular}{ll|cccccccccccc}
        \toprule
          Clipping Value & Accuracy & 0 & 1 & 2 & 3 & 4 & 5 & 6 & 7& 8 & 9 & 10 & 11 \\
        \midrule
            BERT-MRPC (GELU)  &87.75 &   &   &   &  &   &  &  &  &  &  &  & \\
            \hspace{0.5em} 80.0& 87.25& 0.00& 0.00& 0.00& 0.00& 0.00& 0.00& 0.00& 0.00& 0.00& 0.15& 4.33& 0.00\\
            \hspace{0.5em} 60.0& 87.25& 0.00& 0.00& 1.88& 0.00& 0.00& 0.00& 0.00& 0.00& 0.00& 0.31& 4.58& 0.00\\
            \hspace{0.5em} 40.0& 87.01& 0.00& 0.00& 3.76& 0.00& 0.00& 0.00& 0.00& 0.00& 0.00& 2.29& 4.68& 0.00\\
            \hspace{0.5em} 20.0& 87.01& 0.00& 0.00& 3.76& 0.00& 0.00& 0.00& 0.00& 0.00& 0.01& 3.65& 4.79& 0.00\\
            \hspace{0.5em} 10.0& 87.25& 4.64& 1.36& 3.76& 0.00& 1.73& 3.77& 0.00& 0.00& 0.15& 4.53& 4.84& 0.04\\
            \hspace{0.5em} 5.0& 87.25& 49.88& 19.6& 7.11& 7.89& 13.99& 26.68& 34.13& 19.79& 15.1& 5.00& 4.98& 45.9\\
            \hspace{0.5em} 2.0& \bf 84.07& 99.17& 98.87& 98.81& 98.9& 97.62& 97.38& 97.01& 96.39& 94.54& 75.55& 45.77& 92.98\\
            \hspace{0.5em} 1.5& 78.92& 99.96& 99.98& 99.94& 99.94& 99.77& 99.83& 99.77& 99.76& 99.75& 94.72& 78.11& 98.01\\
        \midrule
            BERT-QNLI (MHA-LN)  &91.84 &   &   &   &  &   &  &  &  &  &  &  & \\
            \hspace{0.5em} -60& 91.67& 1.96& 2.68& 0.00& 3.92& 3.92& 3.92& 3.92& 3.92& 3.92& 0.00& 0.00& 0.00\\
            \hspace{0.5em} -55& 91.69& 1.96& 6.21& 1.96& 3.92& 3.92& 3.92& 3.92& 3.92& 3.92& 0.22& 0.00& 0.00\\
            \hspace{0.5em} -50& 91.43& 2.91& 10.69& 1.97& 3.92& 3.92& 3.92& 3.92& 3.92& 3.92& 3.88& 0.06& 0.00\\
            \hspace{0.5em} -45& 91.25& 9.85& 16.15& 7.17& 3.92& 3.92& 3.92& 3.92& 3.92& 3.92& 3.92& 0.45& 0.00\\
            \hspace{0.5em} -40& 90.28& 16.96& 23.13& 13.61& 5.71& 3.92& 3.92& 3.92& 3.92& 4.32& 4.23& 1.46& 0.00\\
            \hspace{0.5em} -35& \bf 85.54& 22.51& 29.36& 23.42& 7.46& 4.50& 3.92& 3.92& 3.92& 5.52& 5.57& 3.21&0.01\\
            \hspace{0.5em} -30& 78.36& 27.48& 36.67& 32.87& 9.39& 7.92& 3.92& 3.92& 4.04& 6.73& 5.86& 6.46&8.19\\
            \hspace{0.5em} -25& 72.73& 34.81& 42.43& 41.88& 23.05& 15.63& 6.55& 3.92& 6.03& 8.16& 5.99& 8.47& 13.84\\
            \hspace{0.5em} -20& 72.52& 41.74& 47.64& 49.76& 37.66& 33.98& 13.51& 6.55& 8.31& 19.08& 6.94& 10.29&40.53\\
        \midrule
      Clipping Value & Accuracy & 0  & 1  & 2  & 3 & 4  & 5 & 0 & 1  & 2 & 3 & 4 & 5  \\
        \midrule
        BART-CoLA (GELU)   & 56.32&  &  &   &  &   & &   &   &  &  &  &   \\
        \hspace{0.5em} 80.0& 56.32& 0.00& 0.00& 0.00& 0.00& 0.00& 0.00& 8.48& 8.60& 0.00& 0.00& 0.00& 0.00\\
        \hspace{0.5em} 60.0& 56.32& 0.00& 0.00& 0.00& 0.00& 0.00& 0.00& 8.60& 8.60& 0.00& 0.00& 0.00& 0.00\\
        \hspace{0.5em} 40.0& 56.32& 0.00& 0.00& 0.00& 0.00& 0.00& 0.00& 8.60& 8.60& 8.60& 0.00& 8.60& 0.00\\
        \hspace{0.5em} 20.0& 56.32& 0.01& 0.00& 0.00& 0.00& 0.00& 0.00& 17.21& 17.21& 8.60& 8.61& 8.61& 0.00\\
        \hspace{0.5em} 10.0& 56.32& 8.60& 4.34& 8.60& 8.60& 0.00& 0.00& 17.21& 17.21& 8.60& 8.61& 8.61& 8.60\\
        \hspace{0.5em} 5.0& 56.58& 9.31& 8.80& 8.83& 8.79& 0.13& 0.43& 20.3& 17.23& 8.74& 8.80& 8.87& 9.29\\
        \hspace{0.5em} 2.0& \bf 54.06& 92.49& 90.98& 78.52& 70.7& 79.58& 62.35& 97.27& 92.14& 74.54& 59.41& 53.88& 42.17\\
        \hspace{0.5em} 1.5& 52.37& 98.98& 98.59& 96.46& 94.45& 96.63& 86.5& 99.88& 99.38& 95.1& 87.58& 84.8& 72.38\\
        \bottomrule
    \end{tabular}
    \end{adjustbox}
    \caption{We evaluate the accuracy directly on dev set with output activation cut by the clipping value. The subsequent columns records the ratio of clipped tokens to all tokens each layer. For BART, we also consider the GELU module in Decoder. Bold numbers show the inflection point of accuracy change.}
    \label{tab_appendix_clipping_impact}
\end{table*}

We provide more evidence of accuracy and token impact by clipping the outputs to different levels in  \autoref{tab_appendix_clipping_impact}.

The first thing is that different outliers have very different importance, where some very large values can be clipped sharply but will not introduce large accuracy degradation, whereas the performance decreases quickly with some being clipped. For example, for the outputs of MHA-LN, clipping them from -60 to -45 seems reliable in the FP model and of course friendly in the quantized one. However, clipping from -40 to -35 will induce about 5\% performance loss. 

Another key point is that those large outliers only belong to several tokens regarding the big divergence of the token range. For example, for values in (-60, -45), the clipped tokens are still 3\% for most of the layers. Thus, finding the clipping range from the token perspective can help to jump over the less important area quickly.

\subsection{Comparisons among Token-Wise Clipping and existing methods}
\label{appendix_twc}
We compare the coarse stage of Token-Wise Clipping with OMSE, percentile, and direct step size learning and argue that ours is more effective \autoref{tab_step_size} and efficient \autoref{tab_step_size_time}.

Our Token-Wise Clipping searches superior clipping ratio towards the final performance and works in a remarkably efficient way (Reasons have been explained in \autoref{subsection_token_wise_clipping}) with about 2 minutes evaluating 30 ratios on GLUE tasks. 

On the contrary, OMSE only minimizes the local quantization error and behaves terribly. For instance, it calculates 40 as the best clipping range for the distribution presented in \autoref{fig_clip_accuracy} while 10 is much better. Also, OMSE runs very slowly even with the fast golden section search.

For the direct step size learning and Percentile methods, though they consider the final loss for the clipping range, they still suffer some problems in the case that the unimportant outliers can cover a large area. Direct step size learning without a good initialization point needs a proper learning rate and much tuning time to achieve the key part.  Take an extreme case as an example. In QAT, step size has been tuned sufficiently but we still notice that the quantized model can be further clipped. Besides, as the Percentile builds a histogram of the activation and searches for the best clipping ratio from the value perspective, it is time-costly to jump over the relatively unimportant outliers.

\begin{table*}[ht]
    \centering
    
    \begin{adjustbox}{max width=\textwidth}
    \renewcommand{\arraystretch}{1.2}
    \begin{tabular}{lccccccccc}
        \toprule
          \multirow{2}{*}{\bf Method} & \textbf{CoLA} & \textbf{MNLI} & \textbf{MRPC} & \textbf{QNLI} & \textbf{QQP} & \textbf{RTE} & \textbf{SST-2} & \textbf{STS-B} \\
        & (Matt.) & (acc m/mm) & (f1/acc) & (acc) & (f1/acc) & (acc) & (acc) & (Pear./Spear.) \\
        \midrule
         RoBERTa (FP)                      & 62.50 & 87.75/87.23 & 93.1/90.44 & 92.68 & 88.78/91.6 & 80.51 & 95.18 & 91.04/90.72 \\
        \midrule
         \hspace{0.5em} OMSE \cite{choukroun2019low}  &  1.81 & 72.89/72.65 & 85.38/78.68 & 76.53 & 85.24/88.94 & 64.26 &91.17  & 80.81/81.99 \\ 
        % \hspace{0.5 em} EasyQuant   & 17.65 & 74.54/74.76 & 82.96/74.02 & \bf 81.97 & 78.56/82.99 & 61.73 & 86.24 & 81.05/81.06 \\
         \hspace{0.5em} Step size learning \cite{bhalgat2020lsq+} & 4.64 & 71.77/73.18 & 85.42/79.17 & 77.28 & 85.19/88.91 & 65.34 & 90.71 & 80.23/81.25 \\
          \hspace{0.5em} Percentile \cite{wu2020integer}  & 20.73 & 72.23/73.68 & 84.83/78.43 & 77.16 & 82.21/87.44 & 62.82 & 88.19 & 79.41/79.64 \\
        %  \hspace{0.5em}   Token-Wise Clipping (Coarse)        &  &  &  &  &  &  &  &  \\
         \hspace{0.5em} Token-Wise Clipping (Coarse Stage)  & \bf 34.95 & \bf 80.56/80.84 & \bf 85.05/79.41 & \bf 79.46 &  \bf 85.96/89.31 & \bf 66.43 &  \bf 91.63 & \bf 82.03/82.45\\
        \midrule
         BERT (FP)                      & 59.60 & 84.94/84.76 & 91.35/87.75 & 91.84 & 87.82/90.91 & 72.56 & 93.35 & 89.70/89.28   \\
        \midrule
         \hspace{0.5em} OMSE   & 35.44 & 74.00/73.30 & 81.54/76.47 & 84.66 & 76.07/82.12 & 64.26 & 86.27 & 85.57/86.05 \\ 
        %  \hspace{0.5 em} EasyQuant \cite{wu2020easyquant} & 44.18 & \bf 78.27/79.24 & 85.12/77.21 & 81.97 & 72.15/79.71 & 62.45 & 85.44 & 82.19/82.03 \\
         \hspace{0.5em} Step size learning  & 35.77 & 74.11/73.76 & 82.95/77.94 & 85.19 & 75.79/81.91 &  64.62 & 87.16 & 85.78/86.47 \\
          \hspace{0.5em} Percentile & 37.32 & 72.40/71.69 & 85.09/79.90 & 79.37 & 72.58/80.19 & 61.73 & 87.27 & 86.38/87.29 \\
         \hspace{0.5em} Token-Wise Clipping (Coarse Stage)  & \bf 47.21 & \bf 77.53/78.01 & \bf 85.40/80.39 & \bf 86.47 & \bf 74.98/83.88 & \bf 64.62 & \bf 91.17 & \bf 86.48/87.06 \\
        \bottomrule
        % \hline
    \end{tabular}
    \end{adjustbox}
    % }
    \caption{Comparisons among existing techniques and the coarse stage of Token-Wise Clipping on 6-bit BERT and RoBERTa models. For the percentile, we search its hyper-parameter in [0.999, 0.9999, 0.99999] and report the best one on dev set. It can be seen that only the coarse stage of our method has surpasses others.}
    \label{tab_step_size}
\end{table*}
\begin{table*}[ht]
    \centering
    
    \begin{adjustbox}{max width=\textwidth}
    \renewcommand{\arraystretch}{1.2}
    \begin{tabular}{cccc}
        \toprule
         \multicolumn{2}{c}{\bf OMSE} &  \bf Percentile & \bf Token-Wise Clipping (Coarse Stage)\\
         \cmidrule(l{2pt}r{2pt}){1-2}  
         \cmidrule(l{2pt}r{2pt}){3-3}  
         \cmidrule(l{2pt}r{2pt}){4-4}  
         Grid Search (30 iterations) & Golden Section Search & search (3 times) & Grid Search (30 iterations)  \\ 
         \midrule
        1754s & 439.29s & 301.49s & 135.73s\\
        \bottomrule
        % \hline
    \end{tabular}
    \end{adjustbox}
    % }
    \caption{The time of activation calibration on 256 samples of each algorithm. As direct step size learning takes OMSE as its initialization, we do not compare the time here.}
    \label{tab_step_size_time}
\end{table*}
\subsection{Supplementary results of QAT}
\label{appendix_qat}
We apply our methods to RoBERTa and BART on quantization-aware training. From \autoref{tab_appendix_qat_glue_results}, on RoBERTa, ours still surpasses LSQ+ by 2.54\% on QNLI, 7.53\% on STS-B. On BART models, we achieve an absolute improvement of 1.73–32.11 points against the best baseline. The outlier suppression framework can be extended to other applications, such as integer-only quantization ~\cite{kim2021bert} as well, which proposes the polynomial approximation of non-linear operations for Transformer-based models.

\section{Supplementary related works}
Quantization algorithms are usually grouped into two categories: (1) Quantization-Aware Training (QAT) and (2) Post-Training Quantization (PTQ). The former fine-tunes the FP model to low-bit and embraces good outcomes with awareness of quantization during training. Apart from learning weight for better performance, \cite{choi2018pact, esser2019learned} propose to learn the quantization parameters. The latter, PTQ, usually conducts fast calibration on the FP model with much less computation and fewer data. \cite{choukroun2019low} transforms quantization to a Minimum Mean Squared Error problem. \cite{wu2020easyquant} alternately optimizes the step size of weight and activation towards the matrix multiplication output.

Recently, quantization has become popular in Transformer-based models. For quantization-aware training, \cite{zafrir2019q8bert} explores 8-bit quantization on BERT-like models. \cite{shen2020q} adopts group-wise quantization and applies mixed-precision quantization based on the Hessian information. \cite{zhang2020ternarybert} investigates various distillation losses on BERT and combines the distillation with quantization. \cite{kim2021bert} approximates the nonlinear function in Transformer architectures to enjoy integer-only inference. \cite{fan2020training} quantizes a different random subset of weights each forward pass during training to decrease quantization noise. Moreover, \cite{tao2022compression} explores underlying difficulties of quantizing generative models. Due to the sequential computation nature of this type of model, they find that word embedding is easier to be homogeneous and devise a token-level contrastive distillation method to combat this obstacle. For post-training quantization, \cite{bondarenko2021understanding} notices the structured outliers in Transformer-based models with the occurrence at a few embedding dims and the special separator token. They point out that the high dynamic ranges will even hurt the 8-bit quantization performance and suggest taking per-embedding-group quantization for this unique challenge.  While they walk around the problem and their method brings extra computation burden, we explore the inducement and clipping impact of these structured outliers and solve them without computation overhead.
\section{Supplementary implementation details}
\label{appendix_implementation_details}
For quantizer details, we insert quantization nodes as \autoref{appendix_quantization_nodes}. We adopt symmetric per-channel quantization on weight and asymmetric per-layer quantization on activation. 

For PTQ experiments, we sample 256 examples as the calibration dataset with batch size set to 32 on GLUE benchmark and SQuAD, 4 for CNN/DailyMail and XSum. For learning in the fine-grained stage of the Token-Wise Clipping, we always tune 3 epochs with learning rate 1e-5 across datasets because the first step already produces good outcomes.

For QAT experiments on the GLUE benchmark, we equip our methods with LSQ+~\cite{bhalgat2020lsq+}. The coarse-grained stage of Token-Wise Clipping is used to initialize quantization parameters, the fine-grained stage is removed because LSQ+ has armed with step size learning. About hyper-parameters, learning rate is searched in \{1e-5, 2e-5, 3e-5, 4e-5, 5e-5\}. Batch size is usually set to 32 unless smaller (8 and 16) ones are also tried on small datasets including CoLA, MRPC, RTE, and STS-B. As for epochs, we follow \cite{bondarenko2021understanding} on BERT (3 epochs for MNLI and QQP, 6 epochs for others), \cite{kim2021bert} on RoBERTa (6 epochs for MNLI and QQP, 12 epochs for others), and take 6 or 12 epochs on BART as well. Other hyper-parameters are inspected and kept fixed across datasets including self-attention dropout rate 0.1, hidden states dropout rate 0.0, weight decay 0.0, and warmup ratio 10\%.
For baseline mechanisms like LSQ+ and PACT, we conduct the above learning rate and batch size search as well for fair comparisons.
\begin{table*}[ht]
    \centering
    
    \begin{adjustbox}{max width=\textwidth}
    \begin{tabular}{lcccccccccc}
        \toprule
          \multirow{2}{*}{\bf Method} & {\bf Bits} &\textbf{CoLA} & \textbf{MNLI} & \textbf{MRPC} & \textbf{QNLI} & \textbf{QQP} & \textbf{RTE} & \textbf{SST-2} & \textbf{STS-B} & \multirow{2}{*}{\bf Avg.} \\
        & (W-E-A) & (Matt.) & (acc m/mm) & (f1/acc) & (acc) & (f1/acc) & (acc) & (acc) & (Pear./Spear.) & \\
        \midrule
        % 8-8-8 & \cite{bondarenko2021understanding} & & & & & &\\
        % \hline

        RoBERTa                                             & 32-32-32  & 62.50 & 87.75/87.23 & 93.1/90.44 & 92.68 & 88.78/91.6 & 80.51 & 95.18 & 91.04/90.72 & 86.40 \\ 
        \midrule
         \hspace{0.5em} Quant-Noise \cite{fan2020training} & PQ & - & 83.60/- & - & - & - & - & - & - & - \\
         \hspace{0.5em} PACT \cite{choi2018pact}        & 4-4-4 & 19.43 & 78.72/79.55 & 81.42/73.04 & 84.55 & 85.14/88.91 & 58.12 & 88.76 & 72.15/72.46 & 70.82 \\
         \hspace{0.5em} LSQ+ \cite{bhalgat2020lsq+}       & 4-4-4 & 24.69 & 83.28/83.24 & 83.17/75.0 & 85.12 & 86.96/90.22 & \bf 58.12 & 89.79 & 78.08/78.41 & 73.36 \\
         \hspace{0.5em} \bf Ours      & 4-4-4 & \bf 37.10 & \bf 84.91/85.2 & \bf 84.60/77.70 & \bf 87.66 & \bf 87.24/90.52 & 57.76 & \bf 90.25 & \bf 85.61/85.33 & \bf 76.67 \\
         \hspace{0.5em} LSQ+(+KD)          & 4-4-4 & 30.33 & 87.17/87.27 & 89.39/85.05 & 91.87 & 88.56/91.48 & 61.73 & 92.20 & 83.18/83.10 & 77.97 \\
         \hspace{0.5em} \bf Ours(+KD)       & 4-4-4 & \bf 48.78 & \bf 87.33/87.16 & \bf 91.92/88.97 & \bf 91.93 & \bf 88.81/91.67 & \bf 66.79 & \bf 92.43 & \bf 88.97/88.76 & \bf 82.09 \\

         \midrule
        BART                         & 32-32-32 & 56.32 & 86.45/86.55 & 91.37/87.5 & 92.31 & 88.34/91.39 & 79.06 & 93.35 & 90.11/89.94 & 84.61 \\
        \midrule
         \hspace{0.5em} PACT         & 4-4-4 & 18.72 & 80.57/80.36 & 87.99/82.60 & 85.52 & 85.09/88.19 & 57.40 & 89.45 & 87.49/87.36 & 73.86 \\
         \hspace{0.5em} LSQ+         & 4-4-4 & 18.12 & 82.41/82.29 & 88.35/83.58 & 87.39 & 86.04/89.64 & 57.40 & 90.48 & 86.89/86.86 & 74.55\\
         \hspace{0.5em} \bf Ours     & 4-4-4 & \bf 50.83 & \bf 84.81/84.57 & \bf 90.94/87.01 & \bf 90.92 & \bf 87.88/90.93 & \bf 73.29 & \bf 92.43 & \bf 89.22/89.02 & \bf 82.46 \\

        \bottomrule
        \end{tabular}
    \end{adjustbox}
    % }
    \caption{Comparison among different QAT strategies with low-bit activation on GLUE benchmark for RoBERTa and BART.}
    \label{tab_appendix_qat_glue_results}
    % \vspace{-8pt}
\end{table*}

% \section{Experimental setting}
% \label{appendix_experimental_setting}
% \textbf{Experimental setting. }

% \section{More experimental results}
% \label{appendix_more_results}
% \textbf{Experimental results. }
\begin{table*}[t!]
% \vspace{-1em}
      \begin{adjustbox}{valign=t,max width=\linewidth}
      \begin{tabular}{lccccc}\toprule 
         Tasks & \multicolumn{3}{c}{GLUE} & XSum \\
        %  \midrule
        \cmidrule(l{2pt}r{2pt}){1-1}
        \cmidrule(l{2pt}r{2pt}){2-4}
        \cmidrule(l{2pt}r{2pt}){5-5}
         Bits(W-E-A) & BERT & RoBERTa  & BART &BART\\ 
         \midrule
         32-32-32 & 417.6 & 475.5 & 534.1 &531.8\\
         %\midrule
         8-8-8  & 104.8 & 119.2 & 134.0 &133.4\\ 
         6-6-6  & 78.7 & 89.5 & 100.6 &100.2\\  
         4-4-8 & 52.6 & 59.8 & 67.3 &67.0\\
         4-4-4 & 52.6 & 59.8 & 67.3 &67.0\\
         2-2-4 & 26.5 & 30.1 & 34.0 &33.8\\
        \bottomrule

    \end{tabular}  
    \end{adjustbox}
    \caption{Model size(MB) of quantized models.}

    \label{tab_model_size}
    % \vspace{-8pt}
\end{table*}

\begin{algorithm}[t!]
    \caption{Token-Wise Clipping}
    \label{algorithm_token_wise_clipping}
    \KwIn{ grid search iteration $K$, model with $L$ layers, number of tokens $T$. }
    \{1. Coarse stage:\}\\
    $loss = \,$INF, $s_0=1.0$\\
    \For{$k=0 $ to $K-1$}{
        $\alpha = 1 - 0.01 * k$\;
        \For{$i=1$ to $L$}{
            layer input $\mX$, token $t$ at embedding $j$ $\mX_{t,j}$\;
            $\vo^{u} = \{\max_j \mX_{1,j},\, \max_j \mX_{2,j},\, ...\,, \,\max_j X_{T,j} \}$\;
            $\vo^{l} = \{\min_j \mX_{1,j},\, \min_j \mX_{2,j},\, ...\,, \,\min_j \mX_{T,j} \}$\;
            $ c^u= quantile (\vo^u, \alpha), \,\,c^l = quantile (\vo^l, \alpha)$\;
            $\mX=clip(\mX, c^l, c^u)$\;
	   }
	   Calculate step size $s$ and quantization loss \autoref{equation_quantization_loss}\;
	   \uIf {$loss > loss_k$} {
            $loss= loss_k, \, s_0 = s$  \;
        }
	  }
	 Find the initialization step size $s_0$\;
	 \{2. Fine-grained stage:\}\\
	 Optimize the $s$ using \autoref{equation_gradient_descent} with \autoref{equation_quantization_loss}\;
	 \textbf{return} Optimized step size $s$ \;
\end{algorithm}

\end{document}